\documentclass[final,3p,times]{elsarticle}

\usepackage{amssymb}
\biboptions{comma,sort&compress}
\usepackage{amsmath}
\usepackage{xcolor}
\usepackage{hyperref}       % hyperlinks
\usepackage{url}            % simple URL typesetting
\usepackage{booktabs}       % professional-quality tables
\usepackage{nicefrac}       % compact symbols for 1/2, etc.
\usepackage{microtype}      % microtypography
\usepackage{makecell,multirow}
\usepackage{caption}
\usepackage{multicol}
\usepackage{graphicx}
\usepackage{subfig}
\usepackage{tabularx}
\usepackage{latexsym}
\usepackage{amstext,amsfonts,amssymb,amsmath}
\usepackage{todonotes}
\usepackage[version=3]{mhchem}
\usepackage{epstopdf}
\AppendGraphicsExtensions{.tif}
\usepackage{nonfloat}
\usepackage{color}
\usepackage{bm}
\usepackage{algorithm}
\usepackage{algpseudocode}
\usepackage{multirow}
\usepackage{tikz}
\usepackage{breqn}
\usepackage{etoolbox}

\DeclareMathOperator*{\argmin}{arg\,min}
\usepackage{tikz}
\usepackage{breqn}
\usepackage{etoolbox}

\newrobustcmd*{\myVtriangle}[2]{\tikz{\filldraw[draw=#1,fill=#2] (0cm,0.2cm) --
(0.2cm,0.2cm) -- (0.1cm,0cm) -- (0cm,0.2cm);}}

\newrobustcmd*{\mythickVtriangle}[2]{\tikz{\filldraw[line width=0.3mm,draw=#1,fill=#2] (0cm,0.2cm) --
(0.2cm,0.2cm) -- (0.1cm,0cm) -- (0cm,0.2cm);}}

\newrobustcmd*{\mythickErrorVtriangle}[2]{\tikz{\filldraw[line width=0.3mm,draw=#1,fill=#2] (-0.05cm,0.05cm) --
(0.05cm,0.05cm) -- (0cm,-0.05cm) -- (-0.05cm,0.05cm);  \draw[draw=#1] (0.0cm, -0.12cm) -- (0.0cm, 0.12cm) ; \draw[draw=#1] (-0.06cm, 0.12cm) -- (0.06cm, 0.12cm); \draw[draw=#1] (-0.06cm, -0.12cm) -- (0.06cm, -0.12cm)    }}

\newrobustcmd*{\mytriangle}[2]{\tikz{\filldraw[draw=#1,fill=#2] (0.0cm,0.0cm) --
(0.2cm,0cm) -- (0.1cm,0.2cm) -- (0cm,0cm);}}

\newrobustcmd*{\mysquare}[2]{\tikz{\draw[draw=#1,fill=#2] (0cm,0cm)
rectangle (0.2cm,0.2cm)}}

\newrobustcmd*{\mythicktriangle}[2]{\tikz{\filldraw[line width=0.3mm,draw=#1,fill=#2] (0.0cm,0cm) --
(0.2cm,0cm) -- (0.1cm,0.2cm) -- (0.0cm,0cm);}}

\newrobustcmd*{\mythicksquare}[2]{\tikz{\draw[line width=0.3mm,draw=#1,fill=#2] (0cm,0cm)
rectangle (0.2cm,0.2cm)}}

\newrobustcmd*{\mybarredtriangle}[2]{\tikz{\draw[draw=#1,fill=#2] (0,0) --
(0.2cm,0) -- (0.1cm,0.2cm) -- (0cm,0cm); \draw[draw=#1] (-0.1cm, 0.07cm) -- (0.3cm, 0.07cm)}}

\newrobustcmd*{\mythickbarredtriangle}[2]{\tikz{\draw[line width=0.3mm,draw=#1,fill=#2] (0,0) --
(0.2cm,0) -- (0.1cm,0.2cm) -- (0cm,0cm); \draw[draw=#1] (-0.1cm, 0.07cm) -- (0.3cm, 0.07cm)}}

\newrobustcmd*{\mybarredsquare}[2]{\tikz{\draw[draw=#1,fill=#2] (0,0)
rectangle (0.2cm,0.2cm); \draw[draw=#1] (-0.1cm, 0.1cm) -- (0.3cm, 0.1cm)}}

\newrobustcmd*{\mythickbarredsquare}[2]{\tikz{\draw[line width=0.3mm,draw=#1,fill=#2] (0,0)
rectangle (0.2cm,0.2cm); \draw[draw=#1] (-0.1cm, 0.1cm) -- (0.3cm, 0.1cm)}}

\newrobustcmd*{\mybarredcircle}[2]{\tikz{\draw[draw=#1,fill=#2] (0,0)
circle (0.1cm); \draw[draw=#1] (-0.2cm, 0.0cm) -- (0.2cm, 0.0cm)}}

\newrobustcmd*{\mythickbarredcircle}[2]{\tikz{\draw[line width=0.3mm,draw=#1,fill=#2] (0,0)
circle (0.1cm); \draw[draw=#1] (-0.2cm, 0.0cm) -- (0.2cm, 0.0cm)}}

\newrobustcmd*{\mythickErrorcircle}[2]{\tikz{\draw[line width=0.3mm,draw=#1,fill=#2] (0,0)
circle (0.06cm); \draw[draw=#1] (0.0cm, -0.12cm) -- (0.0cm, 0.12cm) ;   \draw[draw=#1] (-0.06cm, 0.12cm) -- (0.06cm, 0.12cm); \draw[draw=#1] (-0.06cm, -0.12cm) -- (0.06cm, -0.12cm)    }}

\newrobustcmd*{\mydashedline}[1]{\tikz{\draw[draw=#1] (-0.2cm, 0.2cm) -- (-0.1cm, 0.2cm); \draw[draw=#1] (-0.0cm, 0.2cm) -- (0.1cm, 0.2cm)}}

\newrobustcmd*{\mythickcross}[1]{\tikz{\draw[line width=0.3mm,draw=#1] (0,0) --
(0.2cm,0); \draw[line width=0.3mm,draw=#1] (0.1cm,-0.1cm) -- (0.1cm,0.1cm);}}

\newrobustcmd*{\mybarredcross}[1]{\tikz{\draw[line width=0.3mm,draw=#1] (0,0) --
(0.2cm,0); \draw[line width=0.3mm,draw=#1] (0.1cm,-0.1cm) -- (0.1cm,0.1cm); \draw[draw=#1] (-0.1cm,0) -- (0.3cm,0);}}

\newrobustcmd*{\myline}[1]{\tikz{\draw[draw=#1] (-0.15cm, 0.1cm) -- (0.15cm, 0.1cm);\draw[line width=0.3mm,draw=#1] (-0.0cm, 0.0cm);}}

\newrobustcmd*{\mythickline}[1]{\tikz{\draw[line width=0.3mm,draw=#1] (-0.15cm, 0.1cm) -- (0.15cm, 0.1cm);\draw[line width=0.3mm,draw=#1] (-0.0cm, 0.0cm);}}

\newrobustcmd*{\mythickdashedline}[1]{\tikz{\draw[line width=0.3mm,draw=#1] (-0.2, 0.1cm) -- (-0.1cm, 0.1cm); \draw[line width=0.3mm,draw=#1] (-0.0cm, 0.1cm) -- (0.1cm, 0.1cm); \draw[line width=0.3mm,draw=#1] (-0.0cm, 0.0cm);}}

\newrobustcmd*{\mythickdasheddottedline}[1]{\tikz{\draw[line width=0.3mm,draw=#1] (-0.22, 0.1cm) -- (-0.13cm, 0.1cm); \draw[line width=0.3mm,draw=#1] (-0.085cm, 0.1cm) -- (-0.055cm, 0.1cm); \draw[line width=0.3mm,draw=#1] (-0.01cm, 0.1cm) -- (0.08cm, 0.1cm); \draw[line width=0.3mm,draw=#1] (-0.0cm, 0.0cm);}}

\newrobustcmd*{\mycircle}[2]{\tikz{\draw[draw=#1,fill=#2] (0,0)
circle (0.1cm);}}

\newrobustcmd*{\mythickcircle}[2]{\tikz{\draw[line width=0.3mm,draw=#1,fill=#2] (0,0)
circle (0.1cm);}}

\newrobustcmd*{\mydot}[1]{\tikz{\draw[line width=0.3mm,draw=#1] (0,0)
circle (0.025cm);}}

\journal{Journal}

\begin{document}

\begin{frontmatter}

\title{Adversarial sampling of unknown and high-dimensional conditional distributions}
%\title{Adversarial sampling of unknown conditional high-dimensional distributions with preservation of conditional statistics}
%\title{Exploring unknown high-dimensional spaces while preserving conditional statistics}
\author[fir]{Malik Hassanaly\corref{cor1}}
\ead{malik.hassanaly@nrel.gov}
\author[fir]{Andrew Glaws}
\author[sec]{Karen Stengel}
\author[fir]{Ryan N. King}
\address[fir]{National Renewable Energy Laboratory, 80401 Golden CO, USA}
\address[sec]{University of Colorado Boulder, 80309 Boulder CO, USA}
%\address[sec]{National Renewable Energy Laboratory, 1617 Cole Blvd, Lakewood 80401, USA}
\cortext[cor1]{Corresponding author:}

\begin{abstract}
Many engineering problems require the prediction of realization-to-realization variability or a refined description of modeled quantities. In that case, it is necessary to sample elements from unknown high-dimensional spaces with possibly millions of degrees of freedom. While there exist methods able to sample elements from probability density functions (PDF) with known shapes, several approximations need to be made when the distribution is unknown. In this paper the sampling method, as well as the inference of the underlying distribution, are both handled with a data-driven method known as generative adversarial networks (GAN), which trains two competing neural networks to produce a network that can effectively generate samples from the training set distribution. In practice, it is often necessary to draw samples from conditional distributions. When the conditional variables are continuous, only one (if any) data point corresponding to a particular value of a conditioning variable may be available, which is not sufficient to estimate the conditional distribution. This work handles this problem using an a priori estimation of the conditional moments of a PDF. Two approaches, stochastic estimation, and an external neural network are compared here for computing these moments; however, any preferred method can be used. The algorithm is demonstrated in the case of the deconvolution of a filtered turbulent flow field. It is shown that all the versions of the proposed algorithm effectively sample the target conditional distribution with minimal impact on the quality of the samples compared to state-of-the-art methods. Additionally, the procedure can be used as a metric for the diversity of samples generated by a conditional GAN (cGAN) conditioned with continuous variables.

\end{abstract}

%%Graphical abstract
\begin{graphicalabstract}
\begin{figure}[h!]
\centering
\includegraphics[width=0.99\linewidth]{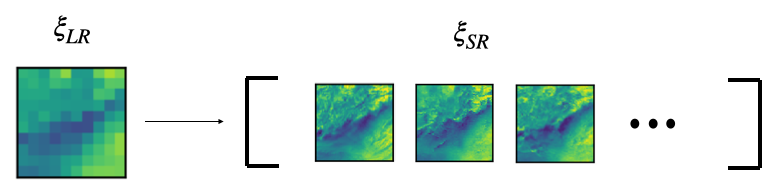}
\caption{Illustration of the outcome of the algorithm. One can generate many high-dimensional samples ($\xi_{SR}$) which are consistent with given low-resolution conditional continuous variables ($\xi_{LR}$).}
\label{fig:Ill}
\end{figure}
\end{graphicalabstract}

%%Research highlights
% 4 to 5 bullet points
% 85 characters at most (with space per bullet point)
%\begin{highlights}
%\item Conditional GANs can be regularized with estimates of conditional moments.
%\item Moments can be estimated with external neural nets or stochastic estimation.
%\item Moments can be used to evaluate the diversity of generated samples.
%% \item Moments can be integrated into the loss function to encourage sample diversity.
%\item The method outperforms state-of-the-art methods for image deconvolution.
%\end{highlights}


\end{frontmatter}

%% \linenumbers
\tableofcontents
%% main text
\section{Introduction}
\label{sec:intro}

\subsection{Problem formulation}
Simulations of physics problems often aim at predicting the average behavior of a system. For example, for design purposes, one needs to predict the average performance of a device under a variety of conditions. However, there exist situations where ensemble averages are not sufficient to achieve the desired prediction. Instead, it may be desirable to access individual realizations of the system. Additionally, in practical applications, one would want to access realizations that correspond to some information already available about the system. In mathematical terms, let $\xi \in \mathbb{R}^N$ be the state of the system, and $g(\xi) \in \mathbb{R}^M$ be observations of the system, where $ 1 \ll M \ll N$. 

The objective of the paper is to draw samples from the conditional distribution  
\begin{equation}
    \label{eq:condDist}
    d \sim \xi \mid g(\xi) = \bar{\xi},
    %d \sim \xi \mid \left\lbrace g(\xi) = \bar{\xi}\right\rbrace,
\end{equation}
for any $\bar{\xi} \in \mathbb{R}^M$ being some partial or low-resolution observation of $\xi$ obtained from $g(\xi)$. For readability, this distribution is also written as $\xi\mid\bar{\xi}$ throughout the paper.

\subsection{Motivation}

In general, since the probability density function (PDF) of $d$ is unknown and high-dimensional, there is no systematic way of sampling it. Nevertheless, this problem is central to a variety of engineering fields. First, it can be used for uncertainty quantification in transient systems. For example, in weather forecasting, one attempts to predict the state of the system at a finite time-horizon using a given set of measurements at the current (and possibly previous) times. The estimation of the atmospheric state given limited observations has been the object of a large body of work \cite{daley1993atmospheric}. However, due to the chaotic dynamics of the atmosphere, any infinitesimal error in the initial state estimate grows exponentially \cite{lorenz1963deterministic} and makes long-term predictions unreliable. To average the chaotic variability and estimate uncertainty in the forecast, it is common to run an ensemble of realizations over the desired finite-time horizon and collect statistics over that ensemble \cite[Chap. 1]{kalnay2003atmospheric}. Generating initial states for these multiple realizations remains an issue for ensemble prediction. It can be formulated as the conditional high-dimensional sampling problem aforementioned. Note that for turbulent fluid flows, this task is even more complex because the underlying distribution may have a relatively small support~\cite{hassanaly2019lyapunov,hassanaly2019ensemble}. One approach in weather forecasting constructs the initial conditions by perturbing a baseline guess along directions that lead to large perturbations growth \cite{toth1997ensemble,buizza1995singular,leutbecher2008ensemble}. While this perturbation-based approach is typically able to capture valuable information for extreme case scenarios \cite{hassanaly2021classification}, the perturbations may not sample the conditional distribution in Eq.~\ref{eq:condDist}. Therefore, the statistics of the ensemble forecast would not reflect the statistics of $d$. Similar considerations apply to other transient processes, whether it refers to the feature of the device \cite{hassanaly2020data} or to the development of an instability \cite{ebi2016experimental,koo2012large}. 

Second, the high-dimensional sampling problem is relevant to evolutionary algorithms that gradually march towards an objective. For instance, in importance splitting algorithms that are used for the estimation of rare event probability, one gradually marches closer and closer to a rare event \cite{morio2014survey,del2005genealogical,cerou2007adaptive}. The path to a rare event is defined as a series of thresholds for a given quantity of interest (QoI). Every time a threshold is attained, the solution is ``cloned" into other candidates while the realizations that did not reach the threshold are discarded. The cloning process is meant to explore the possible paths to a rare event and could also be treated as a high-dimensional conditional sampling process, where $g(\xi)$ is the time history of the quantity of interest. In typical algorithms, the cloning step is done by perturbing the solution that reached a certain threshold \cite{wouters2016rare,hassanaly2019self}. The perturbation procedure is sufficient for certain problems and only in the case where the perturbations are not exceedingly large \cite{wouters2016rare}.

Third, high-dimensional sampling is useful to collect qualitative information about a physics phenomenon. For example, if one observes a system with a coarse temporal resolution, it can be valuable to resolve the path that the system followed between two snapshots. This problem is known as ``in-betweening" or ``temporal super-resolution" \cite{fukami2020machine}. In cases where the governing equation of the system is not fully known or the spatial resolution is coarse, many paths are possible between two snapshots. To obtain these different paths, one can formulate the ``in-betweening" as a conditional sampling problem, where $\bar{\xi}$ are the snapshots obtained at the coarse temporal resolution, and $\xi$ are the temporally super-resolved paths. Another example is the inference of non-resolved fields from resolved ones. This issue is particularly relevant in experiments where measuring some quantities may be difficult. For example, in turbulent combustion experiments, one may want to measure flow velocities as well as species concentrations to understand the effect of the flow field on the flame dynamics. However, measuring both quantities can be challenging from an experimental standpoint. In that case, it may be useful to infer one quantity from the other~\cite{barwey2019using,barwey2020extracting}. Since one maps an input to a higher-dimensional output, several output fields are possible. To obtain these outputs, one can again formulate a conditional sampling problem.

Lastly, the problem targeted in this paper is relevant to the development and testing of coarse-grained models. In coarse-grained approaches, the true solution is projected onto a low-dimensional space that is computationally tractable, and the projected solution is evolved. An example of such an approach is the large eddy simulation (LES) technique for turbulent fluid flows. There, only the large scale motions of the flow are represented, i.e., the solution is projected on the space of large scale fluid motion. When applied to non-linear problems, coarse-grained approaches result in unclosed terms that need to be modeled statistically. In the LES example, the Navier-Stokes  equation for an incompressible flow with homogeneous viscosity is 

\begin{equation}
    \frac{\partial \xi_i}{\partial t} + \frac{\partial \xi_i \xi_j}{\partial x_j } + \frac{\partial p}{\partial x_i} - \nu \frac{\partial^2 \xi_i}{\partial x_j^2} = 0,
\end{equation}

where $\xi_i$ is the $i^{th}$ velocity component, $p$ is the pressure and $\nu$ is the kinematic viscosity. When applying a spatial filter $\overline{(\cdot)}$, the equation becomes 

\begin{equation}
    \frac{\partial \overline{\xi_i}}{\partial t} + \frac{\partial \bar{\xi_i} \bar{\xi_j}}{\partial x_j } + \frac{\partial \overline{p}}{\partial x_i} - \nu \frac{\partial^2 \overline{\xi_i}}{\partial x_j^2} = \frac{\partial (\bar{\xi_i} \bar{\xi_j} - \overline{\xi_i \xi_j})}{\partial x_j },
\end{equation}

where the right-hand-side is unclosed.

Given the true solution $\xi$, a projection operator $g$, and a projected solution $\bar{\xi}$, the ideal closure model is such that~\cite{langford1999optimal,adrian1990stochastic,pope2010self,meneveau1994statistics}

\begin{equation}
    \label{eq:coarseGrained}
    \frac{d \bar{\xi}}{dt} = \left\langle  \left. g \left(\frac{d \xi}{dt}\right) \, \right| g (\xi) = \bar{\xi} \right\rangle,
\end{equation}
where $\langle \cdot \rangle$ denotes an ensemble average and $\frac{d}{dt}$ is the time derivative. The computation of the unclosed terms can be done by directly sampling the high-dimensional space of $\xi$ conditioned on $g(\xi) = \bar{\xi}$. In the context of LES and other coarse-grained approaches \cite{xie2017approximate}, reconstructing $\xi$ is called deconvolution modeling and has received significant attention since the late 90's \cite[Chap. 7]{sagaut2006large}. Apart from a few notable exceptions \cite{germano2009new,adams2004implicit,echekki2001one}, most deconvolution-based closure methods approximate the unfiltered field and directly compute the unclosed terms. One approach is to approximate the filter with a truncated series of filtering operations \cite{stolz1999approximate}. The deconvolved field, supplemented by a stabilization term \cite{stolz2001approximate}, can be used to model unclosed terms in the LES equations. For problems that can be solved in the Fourier space, a spectral extrapolation technique was also proposed \cite{domaradzki1997subgrid} for the amplitude of unresolved wavenumber. The phases of the wavevectors are then arbitrarily set to match that of the unresolved vectors obtained from the non-linear term. An alternative popular approach consists in using a Taylor expansion to approximate the inverse of the filter operation \cite{domingo2015large,domingo2017dns,geurts1997inverse}. A Tikhonov approach has been proposed for various applications \cite{wang2019regularized,wang2019regularizedspray}, where one finds $\xi$ that minimizes $||\bar{\xi} - g(\xi)||^2$. Recently, it was also proposed to reconstruct the unresolved components using an iterative procedure that relies on the existence of an inertial manifold \cite{akram2020priori}. Finally, data-driven methods that take as the input a filtered field and output the deconvolved field have also been explored \cite{maulik2018data,nikolaou2018modelling,yuan2020deconvolutional,brunton2020machine,duraisamy2020machine,fukami2018super,fukami2019super,wang2020physics,stengel2020adversarial}.

While several methods have addressed the deconvolution problem, they suffer from one main deficiency: a single deconvolved field is generated for each convolved field. Instead, a low-dimensional input should map to many high-dimensional outputs, where the variance of the high-dimensional outputs should reflect the uncertainty in the upsampling step. In practice, a one-to-one mapping may be sufficient when the variance of $\xi | g(\xi) = \bar{\xi}$ is low, i.e. when $\bar{\xi}$ almost fully determined the state of the system, and $\xi | g(\xi) = \bar{\xi}$ approaches a delta distribution. In fluid flows, this case may arise for small LES filter sizes, or in systems with low levels of turbulence. The ability to tackle cases where  $\xi | g(\xi) = \bar{\xi}$ does not have low variance could enable using LES with very large filter sizes and for large Reynolds number.

In this paper, an adversarial data-driven approach is proposed to sample conditional high-dimensional distributions. The method leverages generative adversarial networks (GANs) \cite{goodfellow2014generative}, which can theoretically sample from the distribution of any arbitrary dataset. However, GANs are notoriously difficult to train \cite{salimans2016improved,isola2017image}, and the convergence of the training procedure is not guaranteed. Furthermore, conditioning a dataset on the value of continuous variables inevitably decreases the size of the pool of data available. To ensure that the samples generated are diverse enough to span the support of the target conditional distribution, the training procedure is augmented with an a priori estimate of the conditional moments of the conditional distribution. The new procedure is particularly useful in the case where the conditioning variables are continuous. The a priori estimates can be used 1) to evaluate whether the generated samples span the correct distribution, and 2) to regularize the training of the GAN. The rest of the paper is organized as follows. Section~\ref{sec:ganSampling} discusses data-based approaches for sampling high-dimensional distributions and reviews the application of GANs to physics problems. The deficiencies of the training and evaluation procedure of GANs when sampling conditional high-dimensional distributions are highlighted. In Sec.~\ref{sec:method} the algorithm proposed is described. The method is illustrated in Sec.~\ref{sec:srwind} with the deconvolution problem applied to turbulent fluid flow data. It is shown that with the present procedure, many high-quality deconvolved fields can be obtained from a single convolved field. The results obtained with two different a priori estimation of conditional moments are shown and compared to state-of-the-art methods. Conclusions and perspectives are provided in Sec.~\ref{sec:conclusions}.

\section{Sampling unknown distributions with a data-based approach}
\label{sec:ganSampling}

The objective of the paper is to construct an ensemble of realizations $\xi_i$ that are consistent with a given observation $\bar{\xi_i}$, such that $g(\xi_i)=\bar{\xi}_i$. The proposed data-based algorithm to accomplish this assumes that one has access to a pool of realizations randomly sampled for the system of interest. This dataset is then used to sample unseen realizations. 

\subsection{Suitability of Markov Chain Monte Carlo}

In addition to the aforementioned data-based deconvolution approaches~\cite{maulik2018data,nikolaou2018modelling,yuan2020deconvolutional,brunton2020machine,duraisamy2020machine,fukami2018super,fukami2019super,wang2020physics,stengel2020adversarial}, several generic methods have been developed to address the same problem. One popular method is the Markov Chain Monte Carlo (MCMC) method \cite{craiu2014bayesian}. In MCMC, a sequence of samples is constructed according to a rejection rule such that the ensemble of samples matches a target distribution \cite{metropolis1953equation,hastings1970monte,tanner1987calculation}. Because the samples are not independent, a significant proportion of them is not rejected, which is desirable for recovering high-dimensional PDFs. Nevertheless, MCMC requires the knowledge of the shape of the distribution to sample, i.e., the PDF up to a proportionality constant. When one must sample an unknown distribution - as is the case for initial conditions of the atmosphere - the target PDF must be obtained via Bayesian inference. Given the data $\mathcal{D} = \{ \xi_1, ..., \xi_n \}$, where $\xi_i \in \mathbb{R}^N$, one can estimate

\begin{equation}
    P(\xi|\mathcal{D}) \propto P_{likelihood}(\mathcal{D}|\xi) P_{prior}(\xi),
\end{equation}

where $P_{likelihood}(.)$ is the likelihood, and $P_{prior}(.)$ is the prior. However, a reasonable likelihood is not easy to obtain in the case where the variables are correlated, such as for turbulent flow fields. Even in the case where a perfect likelihood could be obtained, a different PDF would have to be estimated for every observation $\bar{\xi}$. Furthermore, if these observations are continuous variables, then only one, if any, realization in the pool of data matches the particular observation. These difficulties make MCMC not suited for the goal of this paper.

\subsection{Generative adversarial networks as a sampling tool}

This work proposes to use GANs to perform the sampling operation. In GANs, a set of training data $\mathcal{D} = \{ \xi_1, ..., \xi_n \}$ and a pair of neural networks are used to generate new samples that span the distribution of the training data. The first neural network is the generator $G$, whose role is to generate these samples. The other neural network is the discriminator $D$, whose role is to decide whether given data is real (i.e., coming from the training data) or fake (i.e., coming from the generator). This network outputs a scalar between 0 and 1, which reflects the probability that the input data is real. The generator and the discriminator compete during training until an equilibrium is reached. In the original version of GANs, the input of the generator is a random vector $z$, whose dimension and distribution may be chosen arbitrarily. New samples can be generated by sampling different values of $z$. Once converged, it can be shown that the generated samples span the same distribution as the training data \cite{goodfellow2014generative}. Given $m$ samples of $G(z)$ and $m$ samples from the training data, the discriminator is rewarded when it distinguishes synthetic samples from true samples. Its adversarial loss is 

\begin{equation}
    \label{eq:discriminatorloss}
    %\mathcal{L}_{adv,D} =   - \mathbb{E}_{\xi} ( log( 1 - D(\xi) ) )  -  \mathbb{E}_{z} ( log( D(G(z)) ) ),
    \mathcal{L}_{adv,D} =   - \mathbb{E}_{z} ( log( 1 - D(G(z)) ) )  -  \mathbb{E}_{\xi} ( log( D(\xi) ) ),
\end{equation}

where $\mathbb{E}_{z}$ denotes the expectation with respect to the distribution of $z$ and $\mathbb{E}_{\xi}$ denotes the expectation with respect to the distribution of $\xi$. The generator is rewarded when it fools the discriminator, and its adversarial loss is 

\begin{equation}
    \label{eq:generatorAdversarialloss}
    \mathcal{L}_{adv,G} = -\mathbb{E}_{z} (log(D(G(z)))).
\end{equation}

In the end, the two networks play the minmax two-player game

\begin{equation}
    \label{eq:minmax}
    \min_G \max_D \mathcal{L}(D,G) = \mathbb{E}_{\xi} \left[ log (D(\xi)) \right] + \mathbb{E}_{z} \left[ log(1-D(G(z))) \right].
\end{equation}

Originally, GANs were successfully applied for image generation techniques \cite{goodfellow2014generative,salimans2016improved,ledig2017photo,isola2017image,karras2020analyzing} but have since been used in the physics community. For example, GANs were used to generate the solution to heat transport equation given arbitrary boundary conditions \cite{farimani2017deep} and to solve stochastic partial differential equations \cite{yang2020physics}. GANs were also able to generate realizations of turbulent flows with realistic spatial statistics despite never enforcing them explicitly \cite{king2018deep}. Compared to other generative models \cite{kingma2013auto,de2019deep}, the ability of GANs to generate high-quality realizations from the high-dimensional distributions underlying the training dataset makes them ideal to tackle the problem of interest here. In particular, the method used in this work leverages the conditional GANs (cGAN) \cite{mirza2014conditional,huang2017stacked} which allows sampling from conditional distributions. For cGANs, the input to the generator is augmented with the value of the conditional variable. 

\subsubsection{Evaluation of GANs and cGANs}

Despite the remarkable abilities of GANs, the samples generated by GANs may not span the entire PDF of the training data \cite{salimans2016improved}. In that case, the GAN is said to have entered a mode collapse, where the distribution of samples wrongly approaches a delta function. This issue has led many engineering applications to use GANs for the generation of single realizations. For the deconvolution problem, GANs have been used to generate a single deconvolved field \cite{stengel2020adversarial,kim2020unsupervised,subramaniam2020turbulence} and sometimes explicitly encourage the generated deconvolved field to exactly match the true deconvolved field \cite{bode2021using}. In this context, evaluating a GAN is essential before deploying it, and several techniques have recently emerged \cite{borji2019pros}. Many available methods rely on inferring some properties about the true distribution of the data to compare it to the generated data. For example, the Parzen window \cite{breuleux2010unlearning} or the coverage metric \cite{tolstikhin2017adagan} approximate the shape of the distribution of the true samples. However, these methods fail if one decides to sample a conditional PDF where the conditional variables are continuous, as is the case in many physics-based applications. For any particular value of the conditioning variable, only one, if any, corresponding data point would be available in the training data, which would not be enough to approximate the true conditional distribution.

Another popular formulation is the Inception score \cite{salimans2016improved} which relies on labeling the images (with discrete labels) using the pre-trained Inception network \cite{szegedy2015going}. While this method could be used in a conditional or unconditional setting, it is not appropriate for physics problems. The Inception score evaluates the diversity of generated images with their labels predicted by the Inception network. These labels are names of objects which images were used to train the network, and may not correspond to the system of interest. Even in the case where one would train a different network to replace the Inception network for the specific system of interest, it is unclear what the output labels should be. Alternatively, the Fr\'{e}chet Inception distance (FID) \cite{heusel2017gans} approximates the moments of the feature distribution and was found to be successful at evaluating the diversity and quality of samples despite only approximating the first two moments. The work reported here shows how to extend the FID to continuous conditioning variables, i.e., how to compute the moments of a conditional PDF with continuous conditioning variables. Besides, both the Inception score and FID cannot disentangle quality and diversity. Here, it is shown how to quantify diversity alone.

\subsubsection{Improving the diversity of samples}

To ensure that the generated samples span the correct PDF (i.e., to combat the mode collapse), several methods have been proposed and can be categorized into four families: 1) increase the gradient of the generator with respect to the noise variable, 2) avoid vanishing gradients of the discriminator far from the training samples, 3) improve the training convergence properties, and 4) match the distribution of the training samples. The first family of methods often tries to increases the diversity of generated samples independently of the noise variables. For example, the similarity between generated samples can be used as a term in the loss function \cite{zhao2016energy,huang2017stacked}. Other approaches explicitly include the gradient of the generator as part of the training loss function either indirectly \cite{zhu2017toward} by ensuring that the generator is invertible, or directly \cite{yang2019diversity,odena2018generator} by estimating the gradient with each minibatch. While these methods have shown promising results in a variety of applications, they do not include a precise quantification of diversity. With the method introduced here, it is possible to estimate a priori the appropriate amount of diversity. Note that in Yang et al.~\cite{yang2019diversity}, the authors attempt to include this information via a tunable parameter $\tau$ but leave open the question of systematically determining $\tau$. Using the method proposed here, it is possible to infer the appropriate value of $\tau$ from the training data. Instead of solely encouraging diversity, the proposed method also informs the cGAN of the spatial distribution of the diversity. For example, in the case of an image of a digit of the MNIST database, pixels at the corner of an image should be less diverse than the ones at the center. The proposed method will convey this information to the network so that diversity is only generated where needed. 

The second family of methods (avoid vanishing gradients for the discriminator) received the most traction from Wasserstein GANs (WGAN) \cite{arjovsky2017wasserstein,gulrajani2017improved}. These methods could be combined with the present work and are left for future work. Similar to the first family of methods, WGANs do not include a quantitative measure of diversity. 
The third family of methods includes training procedures such as the unrollment of the generator \cite{metz2016unrolled}. Again, this technique could be combined with the present work and is left for future work. 
Finally, the fourth family of techniques explicitly attempts to ensure that the diversity of the generated samples matches that of the training samples. For example, in \cite{srivastava2017veegan} the generator is inverted, and it is ensured that the mapping from the training data to the noise variable $z$ matches the prior distribution chosen for $z$. Alternatively, in the mini-batch discrimination technique \cite{salimans2016improved}, the discriminator sees multiple generated samples and compares them to the distribution of the training data to generate extra information and decide whether the generator entered a mode collapse. This family of techniques is appealing in the case of unconditional GANs or cGANs with discrete labels. For continuous conditioning variables, only one sample, if any, is available for a given conditioning value. The method proposed here alleviates the latter issue and could be categorized as part of the fourth family of methods. Using the estimates of conditional moments shown in this paper, an additional term in the loss function can be introduced to ensure that the generated samples span the correct conditional PDF. Similar methods were employed before, albeit in an unconditional setting \cite{wu2020enforcing}.

\section{Method}
\label{sec:method}

The core of the proposed method relies on the approximation of low-order moments of the target conditional distribution. Suppose that the one wants to sample a random variable $\xi \in \mathbb{R}^N$ conditioned on some observed variable $g(\xi)=\bar{\xi} \in \mathbb{R}^{M}$. Encouraging the low-order moments of the true conditional distribution to match the low-order moments of the samples generated helps ensure that the sampled elements span the true conditional PDF. To do this, one first needs to approximate the low-order moments $\mathbb{E}(\xi^p| g(\xi)=\bar{\xi})$ from the true data, where $p$ is an integer corresponding to the desired moment. These quantities are then compared to the sample estimates of the low-order moments $\mathbb{E}(\widehat{\xi}^p| g(\widehat{\xi})=\bar{\xi})$, where $\widehat{\xi} \in \mathbb{R}^N$ denotes the sampled data. 

For ease of notation, distributions of the type $\widehat{\xi}|(g(\widehat{\xi})=\bar{\xi})$ are simplified to $\widehat{\xi}|\bar{\xi}$ throughout the remainder of this paper. Since other metrics that promote diversity were able to give satisfying results with only the first two moments of the target distribution \cite{heusel2017gans,karras2017progressive}, it is chosen to use $p \in \{1,2\}$. Since $\xi$ is a high-dimensional vector, the estimation of the full covariance matrix of $\xi$ may be intractable. Therefore, the variance of $\xi$ is approximated as a matrix made only of its diagonal entries. Mathematically, it can lead the entries of $\xi$ to vary independently of one another. However, the results reported in Sec.~\ref{sec:results} suggest that the discriminator successfully inhibited this unphysical behavior. An interesting extension of this work could consist is investigating the effect of a non-diagonal covariance matrix for the regularization of the cGAN. For turbulent flow applications, a diagonal-by-block covariance matrix may be constructed using the integral length scale of the flow. %To avoid confusion, notations are not changed to reflect this approximation. 

\subsection{Estimate of conditional moments}

The estimate of $\mathbb{E}(\widehat{\xi}^p| \bar{\xi})$ can be easily achieved if one can draw new samples of $\widehat{\xi}|\bar{\xi}$. However, given a fixed training dataset, the estimation of these conditional moments is less straightforward.
When the conditioning variable $\bar{\xi}$ is continuous, it is almost certain that no two training data points share the value of $\bar{\xi}$. In the deconvolution example, the larger the dimension of the convolved field, the more unlikely it is to find two deconvolved sharing the same convolved field. Therefore, the true conditional moments cannot be constructed with a simple Monte Carlo estimator. Instead, one may seek to construct an approximation of the conditional moments at $g(\xi)=\bar{\xi}$ using observations at $g(\xi) \neq \bar{\xi}$.

In order to estimate conditional moments, one can interpret $\mathbb{E}(\xi^p| g(\xi)=\bar{\xi})$ as the function of the random variables $\bar{\xi}$. Estimating the conditional moments can be achieved by finding the function \[
  f\colon \biggl\{\begin{array}{@{}r@{\;}l@{}}
    \mathbb{R}^M &\to \mathbb{R}^N
  ,\\
    \bar{\xi} &\mapsto f(\bar{\xi})
  \end{array}
\] that solves the minimization problem

\begin{equation}
    \label{eq:minimizationProblem}
    \argmin_{f \in \mathcal{F}~s.t.~g(\xi)=\bar{\xi}} ~|| f(\bar{\xi}) - \xi^p ||_2,
\end{equation}

where $||.||_2$ denotes the $L_2$ norm and $\mathcal{F}$ denotes some function space of choice \cite[Chap. 7]{papoulis2002probability}.

Note that once the function $f$ is found, it can be applied to any conditional value $g(\xi)=\bar{\xi}$. In other terms, once $f$ is found, one can estimate the conditional moments for any value of $\bar{\xi}$.

\subsection{Solving the optimization problem}

In principle, any method can be used to obtain the function $f$. Two methods are compared in the paper and their performances are examined in Sec.~\ref{sec:srwind}. 

\subsubsection{Neural network-assisted estimation}

\begin{figure}[ht!]
\centering
\includegraphics[width=0.9\textwidth]{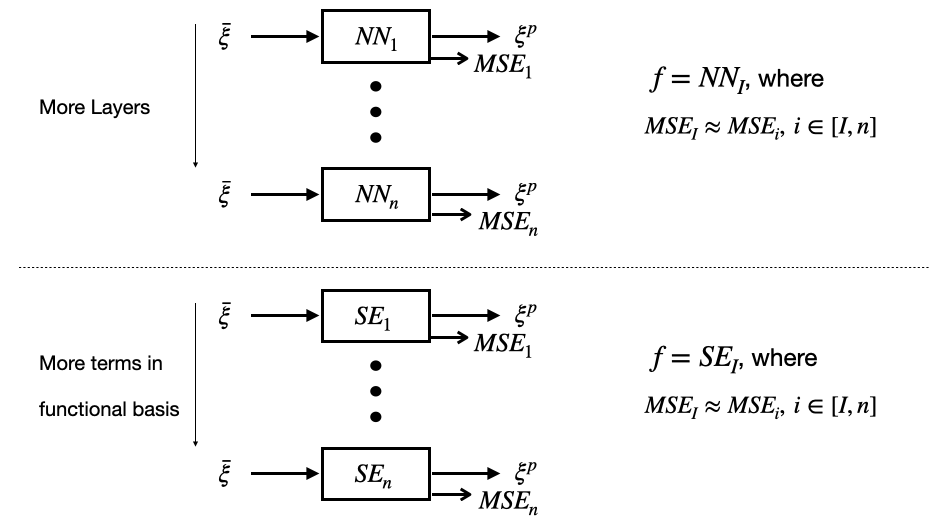}
\caption{Illustration of the procedure adopted for choosing an approximate solution of Eq.~\ref{eq:minimizationProblem}. For the NN-assisted estimation (top), one increases the number of layers until the mean square error (MSE) plateaus. For the stochastic estimation (bottom), one increases the number of terms in the functional basis until the MSE plateaus. The procedure is repeated for each value of $p$.}
\label{fig:NNestimate}
\end{figure}

The first method considered uses a neural network (NN) model with input $\bar{\xi}$ and output $\widetilde{\xi}$. The model is trained in a supervised manner to minimize the mean-square error (MSE) loss between $\widetilde{\xi}$ and $\xi^p$. It can be shown that minimizing the $||\widetilde{\xi} - \xi^p||^2_2$ is equivalent to minimizing the MSE of each variable, separately. The latter approach will be used in Sec.~\ref{sec:stochEst}. To address the minimization problem in Eq.~\ref{eq:minimizationProblem}, different NN are trained with increasing complexity. In practice, one needs to start by choosing some architecture for the NN. A convolutional neural network employs a shared convolutional kernel of weights, making this architecture ideal for spatial data. Varying aspects of the network (e.g., number of layers, kernel size, etc.) changes the size and expressive capabilities of the network. The approximation to $f$ is chosen by increasing these hyperparameters until the final MSE loss stops decreasing and overfitting starts to be apparent (see Fig~\ref{fig:NNestimate}). For each value of $p$, a different NN must be constructed, and the architecture that best fits $p=1$, may not best fit $p=2$.

One can observe the similarity to existing deconvolution procedures that explicitly assign a single deconvolved output as a function of a certain convolved input \cite{maulik2018data,nikolaou2018modelling,fukami2020machine,yuan2020deconvolutional,fukami2018super,fukami2019super,wang2020physics}. In other terms, these methods actually estimate $\mathbb{E}(\xi| g(\xi)=\bar{\xi})$. In the case of low-turbulence intensity or if $M \sim N$, the generated output might look similar to individual realizations of $\xi$, but these methods will generate unreasonably smooth outputs in the case where $M<<N$ and where the turbulence intensity is high.

The advantage of the NN-assisted estimation is the ease of implementation of this method given the availability of open-source libraries that perform the necessary operations. Additionally, it requires no assumptions about the function $f$ that one needs to learn. The disadvantage of the method is that it requires training many different networks and that an arbitrary choice for the baseline architecture of the layers must be made. Another negative aspect of the method is that the NN is used here as a black-box that approximates $f$, and is not easily interpretable.

\subsubsection{Stochastic estimation}
\label{sec:stochEst}
A second method is to choose some $q$-dimensional functional basis to build an expansion of the function $f(\bar{\xi})$. To identify the appropriate number of basis terms in the expansion, one can follow the same principles as the NN-assisted estimation (see Fig.~\ref{fig:NNestimate}). Here, one aims at approximating $f$ as 
\begin{equation}
    \label{eq:funcf}
    f_i(\bar{\xi}) = A_i + \sum_{j=1}^q B_{i,j} b_{i,j}(\bar{\xi}), 
\end{equation} 
where $f_i(\bar{\xi})$ is $i^{th}$ component of $f(\bar{\xi})$, $A_i,B_{i,j}~\in \mathbb{R}$, and $b_{i,j}$ are elements of the functional basis  
\[
  b_{i,j}\colon \biggl\{\begin{array}{@{}r@{\;}l@{}}
    \mathbb{R}^M &\to \mathbb{R}
  ,\\
    \bar{\xi} &\mapsto b_{i,j}(\bar{\xi}).
  \end{array}
\]

The coefficient $A$ can be found by requiring that $\mathbb{E}(f^i(\bar{\xi})) = \mathbb{E}(\xi^p)^i$, where $\mathbb{E}(\xi^p)^i$ is the $i^{th}$ element of $\mathbb{E}(\xi^p)$. The coefficients $B^{i,j}$ can be found by invoking the orthogonality principle \cite{adrian1989approximation,papoulis2002probability} which provides an algebraic expression for the coefficients. The coefficients $B^{i,j}$ can be obtained by solving the linear system 
\begin{equation}
    \label{eq:SE}
    \begin{bmatrix}
    \mathbb{E}(~b_{i,j}(\bar{\xi})b_{i,k}(\bar{\xi})~) 
     \end{bmatrix} 
      \begin{bmatrix}
     B_{i,j}
     \end{bmatrix}
     = 
     \begin{bmatrix}
      \mathbb{E}(~b_{i,j}(\bar{\xi}) \xi_i^p~) 
     \end{bmatrix}  \,.
\end{equation}

The first element in the left handside (LHS) of Eq.~\ref{eq:SE} is a $\mathbb{R}^{q} \times \mathbb{R}^{q}$ matrix, the second element on the LHS of Eq.~\ref{eq:SE} is $\mathbb{R}^{q} \times 1$ vector and the right handside (RHS) is $\mathbb{R}^{q} \times 1$ vector. Note that the coefficients are now expressed as unconditional expectations. One of the functional bases that have proven successful in other contexts is the \textit{stochastic estimation} which uses is a polynomial expansions of $\bar{\xi}$ \cite{adrian1989approximation,adrian1994stochastic,langford1999optimal}. Here, polynomials made with entries of $\bar{\xi}$ are used and the specific expression is provided Tab.~\ref{tab:modelSE}.
To reconstruct $f$, $N$ such systems must be solved - one for each component of $f(\bar{\xi})$ - and the procedure must be redone for each value of $p$. This procedure is therefore expensive but could be accelerated if one can assume some form of homogeneity in the flow field, as is common for turbulent flow problems. In this case, one may not need to solve a linear system for every component of the field. From a computational standpoint, since the stochastic estimation is applied independently for each HR pixel, the process can also be easily parallelized. Similar to the NN-assisted estimation, the elements of the basis must be expanded until $|| f(\bar{\xi}) - \xi^p ||_2$ reaches convergence and stops decreasing. Compared to the NN-assisted estimation, the elements of the basis can be chosen to enforce certain physical constraints and reduce the risk of overfitting and allow interpretability of the result. In Sec.~\ref{sec:srwind} it will be shown that one can use the spatial locality turbulent flow velocity to reduce the size of the functional basis.

Importantly, in either one of the methods, it is not possible to evaluate whether the actual optimal function $f$ has been found for any arbitrary functional space, and it is therefore not possible to know whether $\mathbb{E}(\xi^p| g(\xi)=\bar{\xi})$ is appropriately estimated. However, it is shown below that the samples generated via this technique are quantitatively more diverse than the ones generated by other methods. Furthermore, in Sec.~\ref{sec:srwind} it is shown that the results of the NN-assisted estimation and the stochastic estimation are similar. Therefore, one can expect that a near-optimum of $f$ has been obtained.

\subsection{Evaluation of the diversity of samples}

Given estimates of the conditional moments of the training data, one can train the generated samples to match those moment estimates of the target conditional distribution. It is proposed to estimate the moments of the generated data with a Monte Carlo estimator and to compare the moments with the ones estimated in the a priori study. To quantify diversity, the mismatch between the a priori estimate of the field of standard deviation $\sigma_{\xi}(\xi|\bar{\xi})$ and the field of standard deviation of the generated samples $\sigma_{\widehat{\xi}}(\widehat{\xi}|\bar{\xi})$ is of primary interest. Here, $\sigma_{\xi} (.)$ denotes the standard deviation obtained over the values of $\xi$ and  $\sigma_{\widehat{\xi}} (.)$ denotes the standard deviation obtained over the values of $\widehat{\xi}$.  The norm of the difference between the aforementioned fields, rescaled by the a priori estimate can be interpreted as a percentage of mismatch between the a priori estimate of diversity and the generated diversity. The formal expression of the diversity metric is

\begin{equation}
    \label{eq:divMetric}
    \mathbb{E}_{\bar{\xi}}\left(\frac{|| \sigma_{\xi}(\xi | \bar{\xi}) - \sigma_{\widehat{\xi}}(\widehat{\xi}| \bar{\xi})||_2}{||\sigma_{\xi}(\xi | \bar{\xi}) ||_2}\right),
\end{equation}
where $\mathbb{E}_{\bar{\xi}} (.)$ denotes the expected value taken over the values of $\bar{\xi}$.

\subsection{Integration of the conditional moments in the training procedure}

The main objective of the paper, drawing samples that span the support of high dimensional conditional distribution, can be achieved using the estimated conditional moments. Since the estimation can be done a priori, the moments can be used during the training procedure to encourage the generator to produce diverse samples. For example, it is proposed here to augment the loss function with a ``diversity term" which detects and penalize mode collapse. Inspired by the Fr\'{e}chet Inception Distance (FID) \cite{heusel2017gans}, $\xi$ is assumed to be a multivariate normal with a diagonal covariance matrix, and the loss function is augmented with a diversity loss that is the Fr\'{e}chet distance between the generated samples and the conditional moments of $\xi$. Under these assumptions, the diversity term for every $\bar{\xi}$ can be expressed as \cite{dowson1982frechet}
\begin{dmath}
    \label{eq:divLoss}
    \mathcal{L}_{div}^2 = {|| \mathbb{E}_{\widehat{\xi}}(\widehat{\xi}|\bar{\xi}) - \mathbb{E}_{\xi}(\xi|\bar{\xi} ) ||_2^2} + \\ {\sum_{j=1}^N \mathbb{E}_{\xi}(\xi^2_j|\bar{\xi}) + \mathbb{E}_{\widehat{\xi}}(\widehat{\xi}^2_j|\bar{\xi})  - 2 \sqrt{\mathbb{E}_{\xi}(\xi^2_j|\bar{\xi})\mathbb{E}_{\widehat{\xi}}(\widehat{\xi}^2_j|\bar{\xi}) }}.
\end{dmath}

The approach of moment matching has also been used in an unconditional setting to also fight mode collapse in other physics applications \cite{wu2020enforcing}. The diversity loss is used to augment the loss of the generator, but not of the discriminator, whose role is still to delineate between true and fake samples.

In practice, a mini-batch approach \cite{salimans2016improved} is used to compute the expectations of the type $\mathbb{E}_{\widehat{\xi}}(.)$, where the generator samples $r$ elements conditioned on the same conditional variables. In this work, it was found that $r=25$ gave satisfactory results. In Sec.~\ref{sec:srwind}, alternatives to the proposed diversity loss are explored.

\subsection{Enforcing $g(\xi) = \bar{\xi}$}
\label{sec:enforcingConsistency}

Finally, in order to ensure that the generated samples satisfy $g(\xi) = \bar{\xi}$, one can further augment the generator loss with a ``content loss" which computes the discrepancy between  $g(\xi)$ and $\bar{\xi}$. Here, the content loss is expressed as 

\begin{equation}
    \label{eq:contentLoss}
    \mathcal{L}_{content} = \mathbb{E}_{\bar{\xi}} (\mathbb{E}_{z} (|| g(\widehat{\xi}) - \bar{\xi} ||_2)).
\end{equation}

The same approach to enforcing constraints was adopted in other works \cite{bode2021using,subramaniam2020turbulence,wang2020physics,shah2019encoding}. The effect of the inclusion of physics constraints via a content loss is problem specific and is left for future work that will apply the method described in this paper. In the context of deconvolution, a processing technique has been proposed to exactly enforce that a high-resolution image matches the low-resolution input \cite{sonderby2016amortised}, without using additional loss terms. For the particular problem of deconvolution of box-filtered field, this technique could also be investigated as future work. The constraint $g(\xi) = \bar{\xi}$ of the generated fields is verified a posteriori by computing 

\begin{equation}
    \label{eq:contentMetric}
    \mathbb{E}_{\bar{\xi}} (\frac{1}{||\bar{\xi}||_2}\mathbb{E}_{\widehat{\xi}} (|| g(\widehat{\xi}) - \bar{\xi} ||_2)).
\end{equation}

Equation~\ref{eq:contentMetric} can be interpreted as a relative mismatch between the targeted filtered field, and the filtered generated field. 

To summarize, given a batch size $m$ (chosen here to be 20) which corresponds to the number of independent $\bar{\xi}$ parsed at each step, the discriminator is trained with $r \times m$ true samples and $r \times m$ fake generated samples. When the generator is trained, $m$ true samples of $\bar{\xi}$ are drawn and the generator generates $r$ samples for each $\bar{\xi}$. The discriminator attempts to recognize which ones of the $r\times m$ generated and true samples are fake. To train the generator, the $r \times m$ samples generated are used to compute the adversarial loss $\mathcal{L}_{adv,G}$ (Eq.~\ref{eq:generatorAdversarialloss}), the content loss (Eq.\ref{eq:contentLoss}) and the diversity loss (Eq.~\ref{eq:divLoss}).

With the addition of the content loss and the diversity loss to the adversarial loss, we have arrived at the final form of the generator loss $\mathcal{L}_{G}$

\begin{equation}
    \label{eq:genLoss}
    \mathcal{L}_{G} =  \alpha \mathcal{L}_{content} + \beta \mathcal{L}_{adv,G} + \gamma \mathcal{L}_{div}.
\end{equation}

\begin{figure}[h!]
    \centering
    \includegraphics[width=0.9\textwidth,trim={0cm 0cm 0cm 0cm},clip]{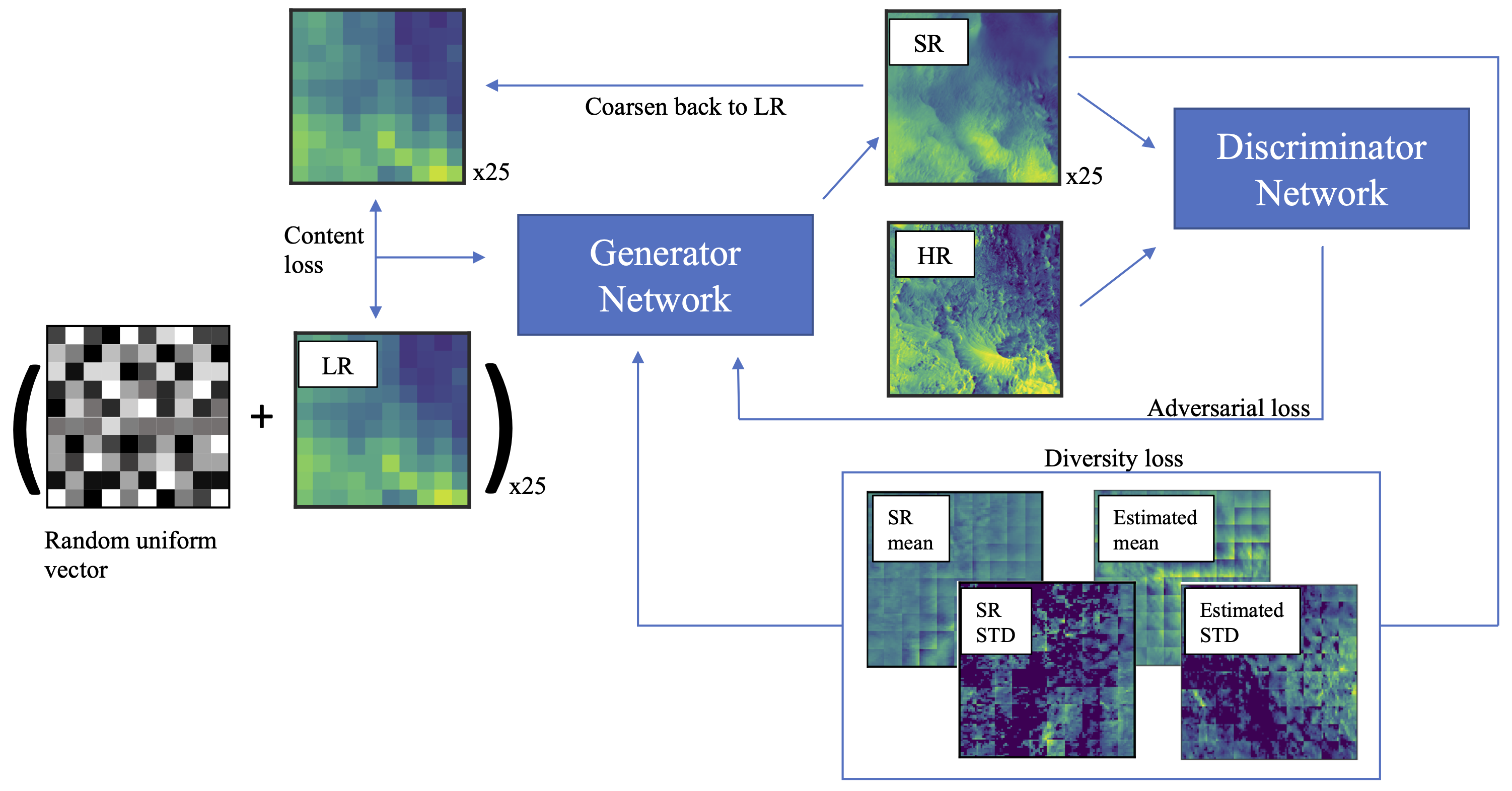}
    \caption{Networks and losses flow. The new diversity loss is computed by sampling 25 SR fields for each LR-HR pair. The moments of the 25 SR fields (SR mean and SR STD in the figure) are compared to the estimated conditional moments (Estimated mean and Estimated STD in the figure).}
    \label{fig:architecture}
\end{figure}

The coefficients $(\alpha, \beta, \gamma)$ scale each term to ensure the proper balance between the three losses (see App.~\ref{app:lossBalance}). For this work, they were chosen to be $\alpha = 1$, $\beta = 0.01$, and $\gamma = 1$. To summarize, the adversarial loss is unchanged compared to traditional cGANs, the content loss verifies that the generated data is consistent with the low-resolution input. The estimated conditional moments are used only in the diversity loss. They ensure that the empirical moments of generated data match with the a priori estimate of the conditional moments. The overall arrangement of the networks and losses is schematically represented in Fig.~\ref{fig:architecture}. 

\section{Illustration: deconvolution of atmospheric data}
\label{sec:srwind}

In this section, the method is applied to the deconvolution of turbulent wind velocity data. The dataset considered is described in Sec.~\ref{sec:dataset}. In Sec.~\ref{sec:estimationWind}, the estimation of the conditional moments is performed, and it is emphasized how convergence of the estimates is assessed. The reader interested in the deconvolution implementation may skip to Sec.~\ref{sec:deconvolutionWind}.

\subsection{Dataset}
\label{sec:dataset}

\begin{figure}[h!]
    \centering
    \includegraphics[width=0.7\textwidth,trim={0cm 0cm 0cm 0cm},clip]{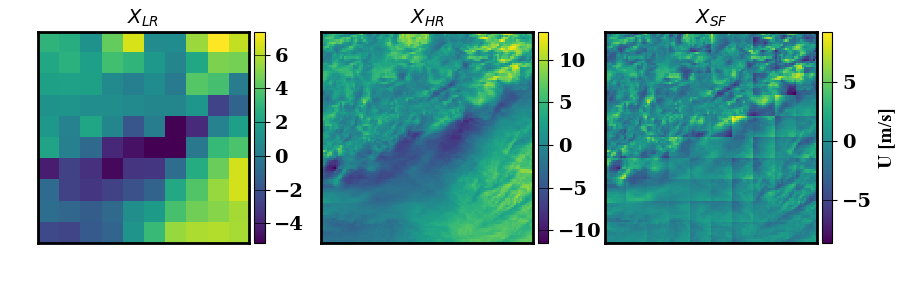}
    \caption{Illustration of the LR, HR and SF velocity. Left: contour of LR easterly wind velocity. Middle: contour of the HR easterly wind velocity. Right: contour of SF easterly wind velocity.}
    \label{fig:illustration}
\end{figure}

The data was obtained from the National Renewable Energy Laboratory's Wind Integration National Database (WIND) Toolkit \cite{draxl2015wind}, which contains easterly and northerly wind speeds at 100m height (a typical wind turbine hub height) over the continental United States for the years 2007-2013. The training dataset is comprised of approximately $38,000$ random snapshots of $100 \times 100$ with a resolution of 10 km/pixel. This data is referred to as the high resolution (HR) dataset. The ground truth HR realizations, derived from NREL's WIND Toolkit data, are noted $\xi_{HR}$. A corresponding low resolution (LR) dataset was generated through a box filter that reduces the data size to  $10 \times 10$ with a resolution of 100 km/pixel. For each LR snapshot $\xi_{LR}$, the goal is to generate a distribution of plausible deconvolved or super-resolved (SR) snapshots $\xi_{SR}$ that exhibit the correct conditional diversity. To simplify notations and visualization, an SR realization $\xi_{SR}$ is decomposed as

\begin{equation}
    \xi_{SR} = \xi_{LR} + \xi_{SF},
\end{equation}

where $\xi_{SR} \in \mathbb{R}^N$ is the SR realization, $\xi_{LR} \in \mathbb{R}^N$ is upsampled to the HR dimension, and $\xi_{SF} \in \mathbb{R}^N$ is the subfilter realization. The objective of the deconvolution is to sample realizations from the distribution 

\begin{equation}
    \label{eq:condDistDeconvol}
    d = (\xi_{SF} | g(\xi_{SF}+\xi_{LR}) = \xi_{LR}),
\end{equation}

where $g$ is the box-filter operator with filter size $\Delta = 10$ pixel composed with a coarsening operation which downselects 1 in every $10$ pixel in every direction. For ease of notations, $d$ is written as $(\xi_{SF} | \xi_{LR})$. A typical snapshot of the dataset along with the LR and the true SF counterpart is shown in Fig.~\ref{fig:illustration}.

\subsection{Estimation of conditional moments}
\label{sec:estimationWind}

In this section, the objective is to estimate the first two moments of the distribution shown in Eq.~\ref{eq:condDistDeconvol}, i.e., to determine $\mathbb{E}(\xi_{SF}|\xi_{LR})$ and $\sigma(\xi_{SF}|\xi_{LR})$, where $\sigma$ denotes the standard deviation of the distribution.

\subsubsection{Neural-network-assisted estimation}
\label{sec:srwindNNEstimation}

First, the neural-network-assisted estimation is implemented. The architecture was chosen to follow the generator architecture of Ledig et al.~\cite{ledig2017photo} with a fully convolutional network with skip connections. The choice of that architecture is motivated by other observations that skip connections helped improve the quality of the generated images \cite{stengel2020adversarial,fukami2018super}. To solve the minimization problem shown in Eq.~\ref{eq:minimizationProblem}, different versions of the NN are trained. Two parameters are varied: the number of residual blocks as well as the number of filters in each convolutional layers. The ratio between the number of filters and the number of residual blocks is arbitrarily held at 2. In total, five NN for each conditional moment are trained with a number of residual blocks in the set $\{ 2,  8, 16, 32 \}$ and the number of filters per convolutional layers in the set $\{ 4, 16, 32, 64\}$. Finer optimization procedures could be used at the cost of training a greater number of NN. Here it will be shown that even suboptimal estimations outperform state-of-the-art methods, and that differences in the conditional moments estimators lead to consistent results. The training set and the validation set are the same as the one presented in Sec.~\ref{sec:dataset}. The neural networks (training and evaluation) were implemented with the Tensorflow 2.0 library \cite{abadi2016tensorflow} and the training of each network took one day on a single graphical processing unit (GPU).

\begin{figure}[h!]
    \centering
    \includegraphics[width=0.45\textwidth,trim={0cm 0cm 0cm 0cm},clip]{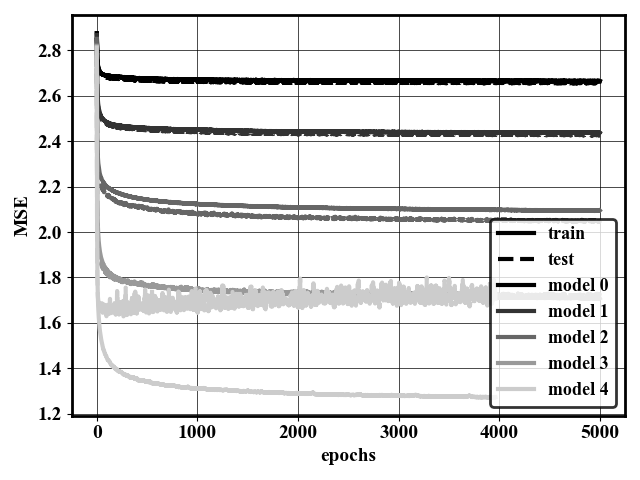}
    \includegraphics[width=0.45\textwidth,trim={0cm 0cm 0cm 0cm},clip]{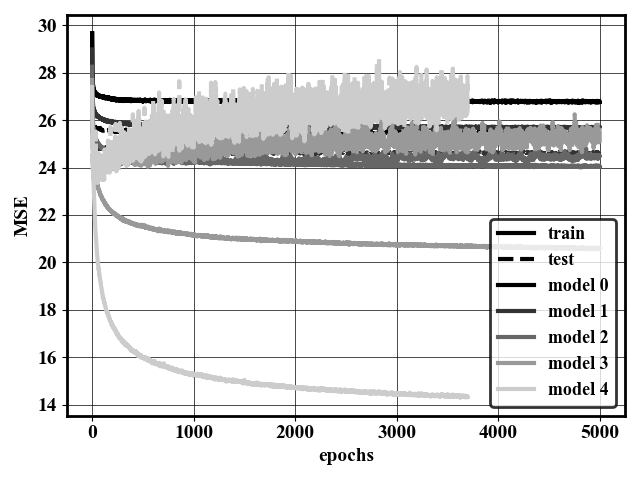}
    \caption{Training history for the NN-assisted estimation. Left: MSE associated for $p=1$. Right: MSE associated with $p=2$. Results with models of increasing complexity (increasingly lighter color) are shown for the training data (\mythickline{black}) and the testing data (\mythickdashedline{black}).}
    \label{fig:trainingNNEstimation}
\end{figure}

The training results are shown in Fig.~\ref{fig:trainingNNEstimation}. As expected, it can be seen that as the number of residual blocks and the number of filters increases, the accuracy of the estimator increases (MSE loss decreases). When the number of trainable parameters in the network becomes too large, overfitting of the training data can be observed and the MSE loss of the testing data stops decreasing. Except in the overfitting cases, convergence with the number of epochs is achieved. For $p=1$ (first conditional moments), Model 3 achieved the best performances without leading to significant overfitting. The same networks were then retrained for $p=2$ (second conditional moment) and in this case, Model 2 achieved the best predictions without overfitting. For $p=2$, the training data was $(\xi_{SF} - \mathbb{E}_{est}(\xi_{SF}|\xi_{LR}))^2$, where $\mathbb{E}_{est}(\xi_{SF}|\xi_{LR})$ is the first moment estimate obtained from Model 3. The MSE loss for both moments and all the models are summarized in Tab.~\ref{tab:trainNNEstimation}.

\begin{table}[t]
\begin{center}
\setlength{\abovecaptionskip}{0pt}
\setlength{\belowcaptionskip}{5pt}
\caption{Training results for the NN-assisted estimations.}
\label{tab:trainNNEstimation}
\begin{tabular}{lcl}
\hline
Model (\# of blocks/\# of filters/\# of trainable parameters)               & MSE $p=1$  &   MSE $p=2$           \\ \hline
Model 0 (2/4/8,468)       & 2.65  & 25.45  \\
Model 1 (8/16/20,392)   & 2.43 & 24.57\\
\textbf{Model 2 (16/32/69,632)} & 2.05  & \textbf{24.55} \\
\textbf{Model 3 (32/64/366,448)}   & \textbf{1.72}  &  24.94  \\
Model 4 (64/128/2,527,568) & 1.70  &  26.27 \\ \hline
\end{tabular}
\end{center}
\vspace{0mm} 
\end{table}

\subsubsection{Stochastic estimation}

Next, the estimation of the conditional moments is performed using stochastic estimation. Here, no assumption of homogeneity is used, which means that the function that approximates the first and second conditional moments differs for each HR pixel. At every pixel, the function to find is chosen to be a polynomial of the conditional data (the LR field). Note that even though the dimensionality of the (not upsampled) LR field is 100 times lower than the HR field, it is still high (200) making intractable the stochastic estimation with a naive polynomial expansion. For instance, a naive polynomial expansion of order 2 -- which was found necessary here (see App.~\ref{app:lseqse})-- would contain 20300 terms which would require inverting a $20300 \times 20300$ matrix. Instead, one can leverage the spatial locality of turbulence to reduce the number of terms in the polynomial expansion.

\begin{figure}[h!]
    \centering
    \includegraphics[width=0.35\textwidth,trim={0cm 0cm 0cm 0cm},clip]{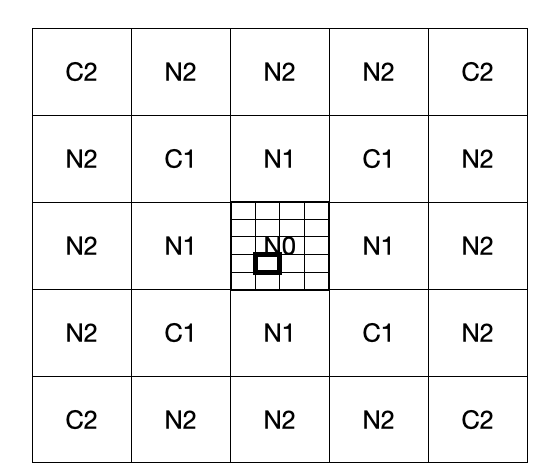}
    \includegraphics[width=0.6\textwidth,trim={0cm 0cm 0cm 0cm},clip]{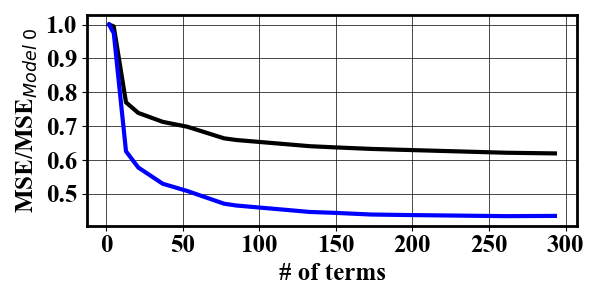}
    \caption{Left: stencil used around each HR pixel to deduce an estimate of conditional moments. The fine grid represents the HR pixels. The coarse grid represents the LR pixels. The moments of the SF velocity of each HR pixel within each LR pixel are obtained as a function of LR pixels in two neighborhood layers (N1-C1 cells and N2-C2 cells). Right: convergence of MSE for $p=1$ (\mythickline{black}) and $p=2$ (\mythickline{blue}) plotted against the number of terms in the polynomial approximation. To ease visualization, the MSE is rescaled against the one obtained with the least amount of terms (Model 0). The  the average absolute value of conditional moment error per pixel of each model compared to the most complex model.}
    \label{fig:trainingSE}
\end{figure}

A stencil composed of the $5 \times 5$ grid of the conditional variable (the LR image) is used to perform the stochastic estimation at each HR pixel. The stencil is centered on the LR pixel $N0$ that contains the HR pixel (see Fig.~\ref{fig:trainingSE}). A Neumann boundary condition (zero-gradient) is used for the boundary pixels. The stochastic estimation is performed with 15 different models for which the size of the functional basis increases. The polynomials used in each model along with the MSE error for each moment are shown in Tab.~\ref{tab:modelSE}. In the table, the einstein notation is used and $U^{(j)}_{XX,i}$ denotes the $j^{th}$ velocity component of the $i^{th}$ $XX$ LR pixel of the stencil, where $XX \in \{N0,N1,C1,N2,C2\}$ and $j \in \{ 1,2 \}$. The notation $U^{(j)}_{X1, i, adj}$ refers to the LR pixel adjacent to the $X1$ LR pixel within the immediate neighborhood of $N0$ in the clockwise direction. The notation $U^{(j)}_{X1, i, opp}$  refers to the LR pixel symmetric to the $X1$ LR pixel with respect to the $N0$ pixel. It can be seen in Fig.~\ref{fig:trainingSE} (right) that the MSE error significantly drops for both moments when the velocities of the immediate neighboring cells are included as part of the functional basis (Model 3 and Model 4). This suggests that the conditional moments of the subfilter component depend on the spatial gradients at the LR resolution. This observation is unsurprising as many subgrid-scale models (SGS) have been successful while using spatial gradients of the filtered field \cite{pope2001turbulent}. Such sanity checks can only be performed with the stochastic estimation since the polynomial basis is more easily interpretable than with the NN-assisted estimation.

\begin{table}[t]
\begin{center}
\setlength{\abovecaptionskip}{0pt}
\setlength{\belowcaptionskip}{5pt}
\caption{List of stochastic estimation models used}
\label{tab:modelSE}
\begin{tabular}{llcc}
\hline
Model (number of terms)  & Terms    &   MSE $p=1$ & MSE $p=2$        \\ \hline
Model 0  (2 terms) &  \{$U^{(i)}_{N0}$\} & 2.78 & 73.83   \\
Model 1  (4 terms) &  Model 0 + \{${U^{(i)}_{N0}}^2$\} & 2.77 & 72.54    \\
Model 2  (5 terms) &  Model 1 + \{$U^{(1)}_{N0} U^{(2)}_{N0}$\} & 2.76 & 72.03  \\
Model 3  (13 terms) &  Model 2 + \{$U^{(j)}_{N1,i}$\}  &  2.14 &  46.2 \\
Model 4  (21 terms) &  Model 3 + \{$U^{(j)}_{C1,i}$\}  & 2.05 &  42.66 \\
Model 5  (37 terms) &  Model 4 + \{$U^{(i)}_{N0}U^{(j)}_{N1,i}$\}  &  1.98 &  39.11 \\
Model 6  (53 terms) &  Model 5 + \{$U^{(i)}_{N0}U^{(j)}_{C1,i}$\}  & 1.94 &   37.54 \\
Model 7  (77 terms) &  Model 6 + \{$U^{(j)}_{N2,i}$\} & 1.84  &  34.77  \\
Model 8  (85 terms) &  Model 7 + \{$U^{(j)}_{C2,i}$\}  & 1.83  &  34.37  \\
Model 9  (133 terms) &  Model 8 + \{$U^{(j)}_{N0}U^{(k)}_{N2,i}$\} & 1.78 & 32.97 \\
Model 10  (149 terms) &  Model 9 + \{$U^{(j)}_{N0}U^{(k)}_{C2,i}$\} & 1.77 & 32.78 \\
Model 11  (173 terms) &  Model 10 + \{$U^{(j)}_{N0}U^{(k)}_{N1,i}U^{(l)}_{N1,i}$\} & 1.76 &  32.41 \\
Model 12  (197 terms) &  Model 11 + \{$U^{(j)}_{N0}U^{(k)}_{C1,i}U^{(l)}_{C1,i}$\}  & 1.75  & 32.30 \\
\textbf{Model 13  (261 terms)} &  \textbf{Model 12 + \{}$\boldsymbol{U^{(j)}_{N0}U^{(k)}_{N1,i}U^{(l)}_{N1,i,adj},U^{(j)}_{N0}U^{(k)}_{C1,i}U^{(l)}_{C1,i,adj}}$\textbf{\}} & \textbf{1.72} & \textbf{32.06} \\
Model 14  (293 terms) &  Model 13 + \{$U^{(j)}_{N0}U^{(k)}_{N1,i}U^{(l)}_{N1,i,opp}$, $U^{(j)}_{N0}U^{(k)}_{C1,i}U^{(l)}_{C1,i,opp}$ \} & 1.72 & 32.11 \\

\hline

\end{tabular}
\end{center}
\vspace{0mm} 
\end{table}

For models more complex than Model 13, it can be observed that the MSE for $p=2$ starts to increase, which suggests that Model 13 can be considered at the optimum of Eq.~\ref{eq:funcf}. The MSE for $p=1$ closely match that obtained with the NN-assisted estimation, however the MSE for $p=2$ is significantly larger (see Tab.~\ref{tab:trainNNEstimation} for comparison). This difference between both approaches is used hereafter to assess the effect of a suboptimal choice for the second conditional moments. The deconvolution will also be done with Model 7 to assess the effect of a suboptimal choice for the first and second moments.

\begin{figure}[h!]
    \centering
    \includegraphics[width=0.7\textwidth,trim={2.5cm 0cm 5cm 0cm},clip]{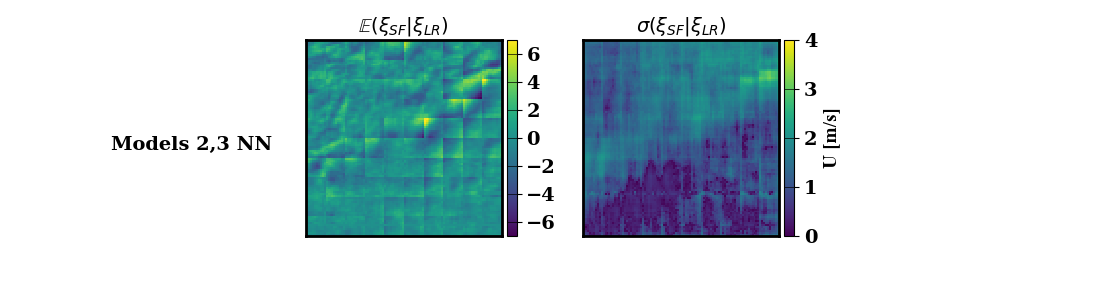}
    \includegraphics[width=0.7\textwidth,trim={2.5cm 0cm 5cm 0cm},clip]{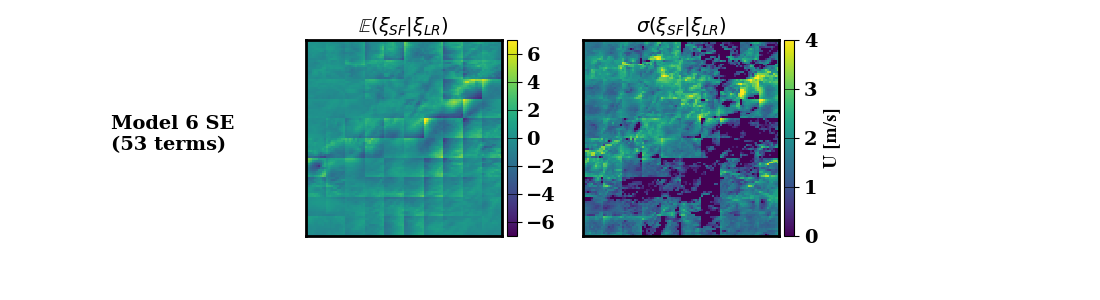}
    \includegraphics[width=0.7\textwidth,trim={2.5cm 0cm 5cm 0cm},clip]{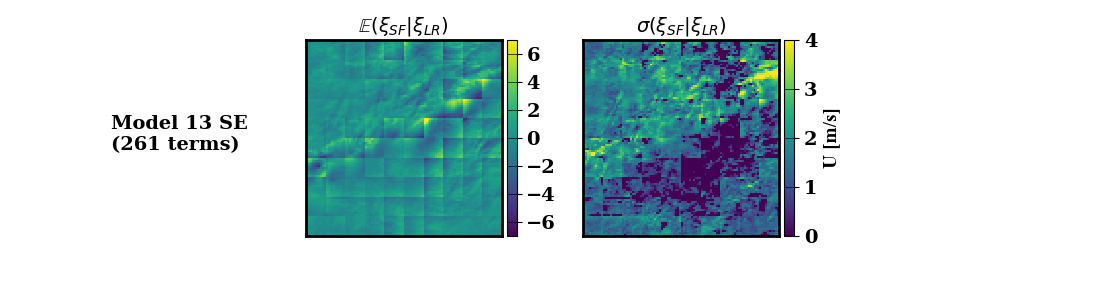}
    \caption{Contours of conditional moments of the SF velocity component for the same snapshot as Fig.~\ref{fig:illustration}. Top: NN-assisted estimation. Model 2 (see Tab.~\ref{tab:trainNNEstimation}) is used for $p=1$ (left) and Model 3 (see Tab.~\ref{tab:trainNNEstimation}) is used for $p=2$ (right). Middle: stochastic estimation. Model 6 (see Tab.~\ref{tab:modelSE}) is used for $p=1$ (left) and $p=2$. Bottom: stochastic estimation. Model 13 (see Tab.~\ref{tab:modelSE}) is used for $p=1$ (left) and $p=2$ (right).}
    \label{fig:illustrationStats}
\end{figure}

The conditional moments obtained with stochastic estimation using a 53 terms model and a 261 terms model are shown in Fig.~\ref{fig:illustrationStats}. The moments correspond to the same snapshot as the one shown in Fig.~\ref{fig:illustration}. Visually, all moments exhibit similar features, despite leading to significant differences in the MSE. The conditional mean of $\xi_{SF}$ is reasonably close to the true $\xi_{SF}$, and the SF energy is largest near shear layers, i.e., regions where turbulence intensity may be large. Likewise the conditional standard deviation of $\xi_{SF}$ is the largest in regions of sharp gradients, which corresponds to the physical intuition. For the same LR field, there exist many SF configurations in high-turbulence intensity regions. The moments of $\xi_{SF}$ also exhibit a jagged structure at the edge of the LR pixels. This structure is not a numerical artifact but is due to the nature of the filter used. The patterns observed are in agreement with the contour of the true $\xi_{SF}$ (Fig.~\ref{fig:illustration} right). 

\subsection{Deconvolution network}
\label{sec:deconvolutionWind}

The deconvolution (or super-resolution) network architecture is based on the approach from Stengel, et al. \cite{stengel2020adversarial} that is capable of performing single-field super-resolution of wind and solar climate data. The generator network $G(\cdot)$ is fully convolutional and uses $3 \times 3$ convolutional kernels with ReLU activations. A preprocessing layer expands the LR input tensor (not upsampled) with two input channels (corresponding to the easterly and northerly wind velocities) to 56 channels and appends a random tensor $z$ (drawn uniformly from the interval $[-1, 1]$) resulting in a tensor with 64 data channels. Next, sixteen residual blocks with skip connections process and prepare the data to be enhanced. Super-resolution blocks increase the spatial resolution data using depth-to-space steps. The discriminator network $D(\cdot)$ is comprised of eight convolutional layers with leaky ReLU activations and two fully connected layers.

The generator network loss function contains three terms:  (i) a content loss, (ii) an adversarial loss, and (iii) a diversity loss as shown in Eq.~\ref{eq:genLoss}.

Following the method outlined in Stengel et al.\cite{stengel2020adversarial}, a balance is maintained between the performances of the generator and the discriminator. If the generator (respectively the discriminator) outperforms the discriminator (respectively the generator) - performance is measured with the value of the adversarial loss function - multiple training iterations are performed for the discriminator (respectively the generator) without updating the generator (respectively the discriminator). 

The implementation of the super-resolution network and the codes used for the moment estimation are made available in a companion repository (\url{http://github.com/NREL/diversity_SR}).

\subsection{Alternative diversity losses}

The proposed diversity loss (Eq.~\ref{eq:divLoss}) is compared to other approaches that also leverage an additional loss term to enforce conditional statistics.

First, the most naive diversity loss (referred to as ``Generator Similarity") is implemented. For each $\xi_{LR}$, one also draws a minibatch of $\xi_{SR}$ fields and penalizes low variance of the minibatch. This is an attempt to generate as much diversity as possible, irrespective of the spatial distribution of diversity or the amount of diversity already present. This approach is also presumably sensitive to the batch of data. If in the batch of data, the fields have a relatively low turbulence intensity, it may not be possible to generate diverse fields. In turn, if the fields have a relatively high turbulence intensity, the generated data should be more diverse. However, this loss will equally penalize the lack of diversity by the same amount. The Generator Similarity loss is expressed as:

\begin{equation}
    \mathcal{L}_{div,GS} = \mathbb{E}_{\xi_{LR}}\left[ - \sigma(\xi_{SR}|\xi_{LR})\right].
\end{equation}

For the Generator Similarity technique, it was found that $\gamma>0.01$ in Eq.~\ref{eq:genLoss} led to instabilities during training. Therefore, $\gamma$ was set to $0.01$ to promote as much diversity as possible.

Second, the approach of Yang et al.~\cite{yang2019diversity} (referred to as ``DSGAN") is implemented. It attempts to increase the gradient of the generator with respect to the noise variable, and takes the form:

\begin{equation}
    \mathcal{L}_{div,DSGAN} = -\mathbb{E}_{\xi_{LR}} \left[ \mathbb{E}_{z_1,z_2} \left[ \min \left( \frac{|| \xi_{SR}(\xi_{LR},z_1) - \xi_{SR}(\xi_{LR},z_2) ||}{||z_1-z_2||},\tau \right) \right] \right],
\end{equation}

where $z_1$ and $z_2$ are two different noise variable values, $\tau$ is a tunable parameter that limits the amount of diversity. In Yang et al.~\cite{yang2019diversity} $\tau$ is a hyperparameter that should be set a priori. Here, $\tau$ is dynamically set to $||\sigma(X_{SF}|X_{LR})||_2/||\sigma(z)||_2$. Therefore, it is attempted to set a level of diversity that is consistent with Eq.~\ref{eq:divLoss}, but that does not take into account the spatial distribution of diversity. Consistently with Yang et al.~\cite{yang2019diversity}, $\gamma =8$. Similar to the Generator Similarity, it was attempted to increase the value of $\gamma$. While diversity increased, the quality of the samples significantly decreased as the generated samples exhibited strong small scale fluctuations.   

\subsection{Alternative deconvolution methods}

For comparison to other popular deconvolution methods used for turbulent flows, the iterative approximate deconvolution method (ADM) \cite{stolz1999approximate,stolz2001approximate,van1931einfluss} and the Taylor series-based deconvolution method \cite{domingo2015large,domingo2017dns} are implemented. Note that ADM relies on the iterative convolution of the convolved fields to reconstruct an approximate of the deconvolved field. Formally, ADM the deconvolved field is approximated as the following truncated sum  

\begin{equation}
    \xi \approx \bar{\xi} + \sum_{i=1}^n (I-g)^i \bar{\xi}, 
\end{equation}

where $g(\xi) = \bar{\xi}$, $g$ is the filter composed with the coarsening operator, and $n$ is the number of terms in the truncated sum. Consistently with Stolz et al.~\cite{stolz1999approximate}, $n=5$ is chosen for the iterative procedure. In the deconvolution problem tackled here, since $g(g(\xi)) = g(\xi)$, the ADM procedure is unsuccessful. The dependence of the ADM procedure with the type of filter used has been investigated elsewhere \cite{san2015posteriori}. For the sole purpose of comparison with ADM, an alternative filtered field obtained with a Gaussian filter is constructed, and no coarsening is applied. The Gaussian filter about the point $(x_0,y_0)$ is  

\begin{equation}
    G(x-x_0,y-y_0) = \frac{6}{\pi \Delta^2} exp(-\frac{6 || x - x_0 ||_2^2 }{\Delta^2}) exp(-\frac{6 || y - y_0 ||_2^2 }{\Delta^2}) , 
\end{equation}

where $\Delta$ is the filter size.

The Taylor expansion-based deconvolution is also implemented in the case of a Gaussian filter. The deconvolved field can be expressed as a truncated sum. The derivation of such approaches has been described elsewhere \cite[Chap. 2 and Chap. 7]{sagaut2006large}. Here the sum is truncated after the first term and leads

\begin{equation}
    \xi \approx \bar{\xi} - \frac{\Delta^{2}}{24} \nabla ^{2} \bar{\xi}.
\end{equation}

For both deconvolution methods, a single deconvolved field can be generated from a single convolved field. Therefore the diversity of the deconvolved fields cannot be compared to the proposed method. For the ADM and the Taylor-based deconvolution methods, only the quality of the samples will be assessed in Sec.~\ref{sec:results}.

\subsection{Results}
\label{sec:results}

\subsubsection{Qualitative assessment of sample quality}

\begin{figure}[h!]
    \centering
    \includegraphics[width=0.99\textwidth,trim={0cm 0cm 0cm 0cm},clip]{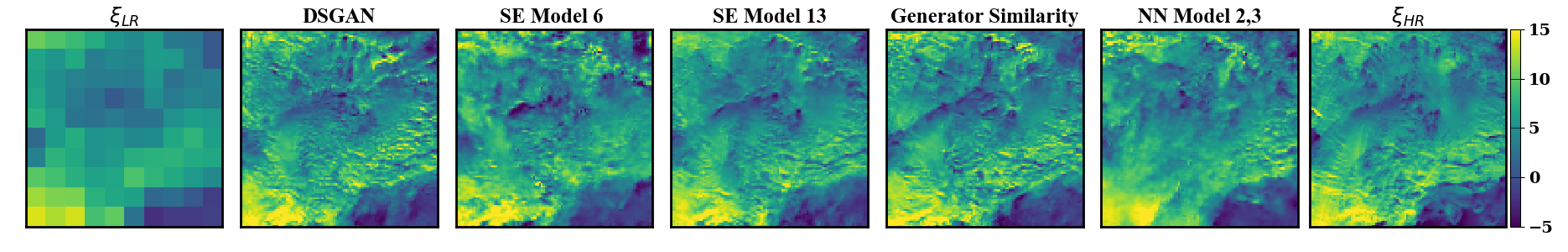}
    \includegraphics[width=0.6\textwidth,trim={0cm 0cm 0cm 0cm},clip]{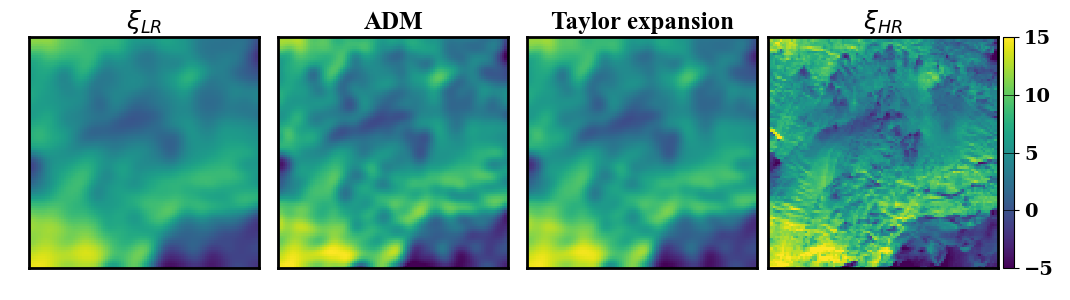}
    \caption{Top: contour of easterly velocity. From left to right: (i) LR field, (ii) field generated with DSGAN, (iii) field generated with the proposed method using SE performed with Model 6 (see Tab.~\ref{tab:modelSE} for the conditional moments, (iv) field generated with the proposed method using SE performed with Model 13 (see Tab.~\ref{tab:modelSE} for the conditional moments, field generated with the Generator Similarity loss, (v) field generated with the proposed method using NN with Model 2 (see Tab.~\ref{tab:trainNNEstimation} for $p=1$ and Model 3 for $p=2$, (vi) the true HR field.
    Bottom: contour of easterly velocity. From left to right: (i) HR field filtered with a Gaussian filter, (ii) field generated with iterative deconvolution (ADM), (iii) field generated with a Taylor expansion approximated deconvolution, (iv) the true HR field. proposed method using SE performed with Model 6 (see Tab.~\ref{tab:modelSE} for the conditional moments, (iv) field generated with the proposed method using SE performed with Model 13 (see Tab.~\ref{tab:modelSE} for the conditional moments, field generated with the Generator Similarity loss, (v) field generated with the proposed method using NN with Model 2 (see Tab.~\ref{tab:trainNNEstimation} for $p=1$ and Model 3 for $p=2$, (vi) the true HR field.}
    \label{fig:illustrationSamples}
\end{figure}

The quality of the generated samples is first assessed by visual examination of the generated samples for one LR field of the validation dataset. Figure~\ref{fig:illustrationSamples} shows an example of LR, HR, and deconvolved (or SR) snapshots obtained with different methods. First, by comparing GAN based approaches (top of Fig.~\ref{fig:illustrationSamples}) to inverse filter approximations (bottom of Fig.~\ref{fig:illustrationSamples}), GANs can generate higher fidelity samples and produce high-gradients in the SR fields. By contrast, the ADM and Taylor expansion approaches tend to smear large gradients. Samples generated by all GAN-based approaches are indistinguishable from the HR fields.

\begin{figure}[h!]
    \centering
    \includegraphics[width=0.485\textwidth,trim={0cm 0cm 0cm 0cm},clip]{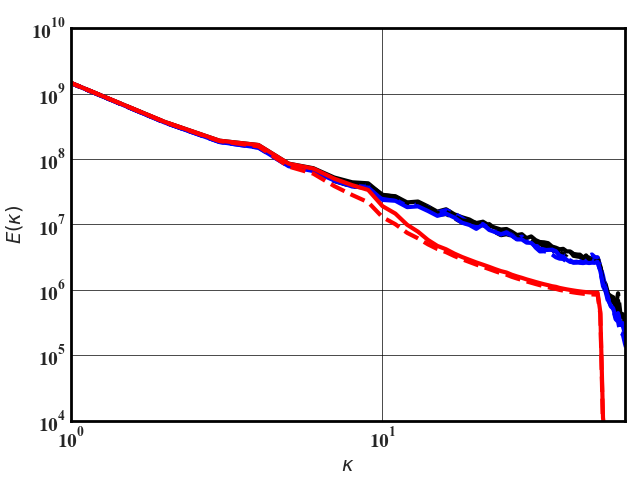}
    \includegraphics[width=0.485\textwidth,trim={0cm 0cm 0cm 0cm},clip]{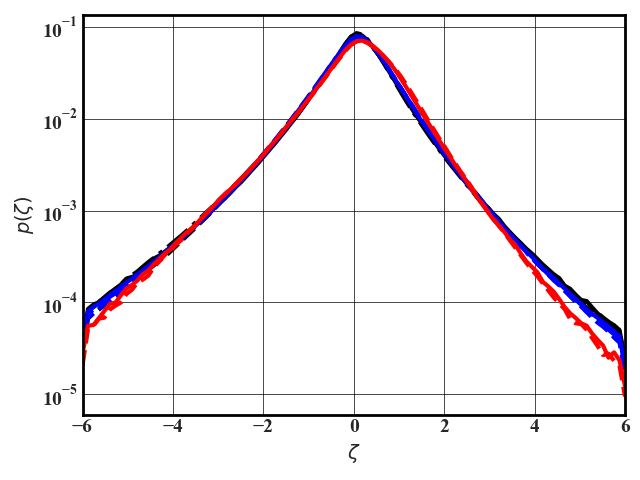}
    \includegraphics[width=0.485\textwidth,trim={0cm 0cm 0cm 0cm},clip]{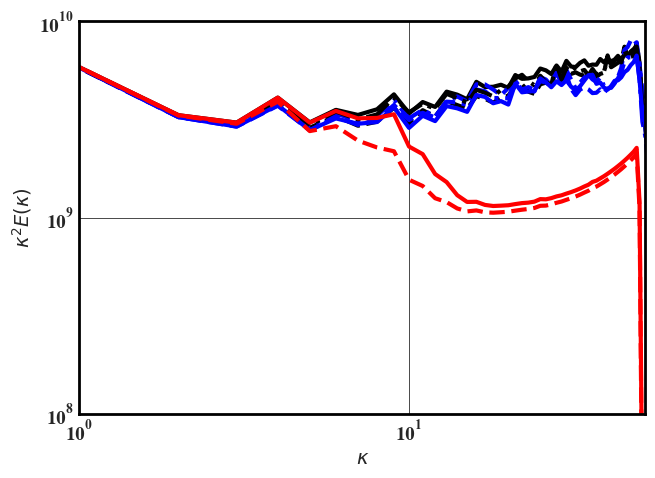}
    \includegraphics[width=0.485\textwidth,trim={0cm 0cm 0cm 0cm},clip]{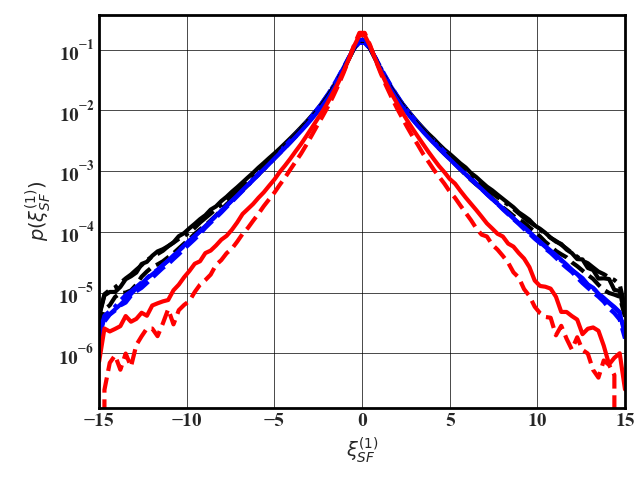}
    \caption{Turbulent statistics of DSGAN (\mythickdashedline{black}), SE with Model 6 (\mythickdashedline{blue}), SE with Model 13 (\mythickdasheddottedline{blue}), Generator Similarity (\mythickdasheddottedline{black}), NN-assisted estimation with Model 2 and 3 (\mythickline{blue}), ADM (\mythickline{red}), Taylor expansion deconvolution (\mythickdashedline{red}), and ground truth data (\mythickline{black}). Top left: turbulent energy spectrum. Top right: PDF of the longitudinal velocity gradient. Bottom left: energy dissipation spectrum. Bottom right: PDF of the subfilter easterly velocity.}
    \label{fig:turbStats}
\end{figure}

The statistics of the velocity fields plotted in Fig.~\ref{fig:turbStats} lead to similar conclusions. The statistics were obtained from the validation dataset. While most GAN-based approaches give an energy spectrum indistinguishable from the HR energy spectrum (left of Fig.~\ref{fig:turbStats}), the ADM and Taylor expansion approaches underestimate the energy of the small scales. The PDF of the longitudinal velocity gradient $\zeta = \frac{\partial U}{\partial x} / \langle (\frac{\partial U}{\partial x})^2 \rangle^{1/2}$, where $U$ is the easterly velocity, $x$ is the west-east direction, and $\langle . \rangle$ denotes the spatial averaging are shown in Fig.~\ref{fig:turbStats} (right). All methods are reasonably close to the statistics of gradients of the training data. Overall, the new diversity loss does not appear to impact the quality of the generated samples. For the ADM and Taylor expansion methods, the probability of large gradients is slightly underestimated which confirms the visual observation. To better distinguish between the different GAN models, the dissipation spectrum (bottom left) and the PDF of the subfilter easterly velocity (bottom right) are also shown. While all GAN models reasonably approximate the HR, compared to DSGAN and Generator similarity, the models introduced tend to slightly underestimate the dissipation spectrum and the tails of the subfilter PDF. This observation can be explained by the fact that the quality of the samples slightly decreased in favor of added diversity. It is in line with the results presented in Tab.~\ref{tab:results}.

\subsubsection{Qualitative assessment of sample diversity}

\begin{figure}[h!]
    \centering
    \includegraphics[width=0.97\textwidth,trim={12cm 0cm 15cm 0cm},clip]{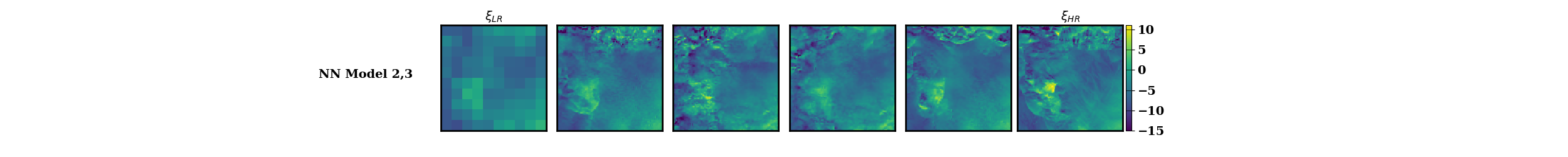}
    \includegraphics[width=0.97\textwidth,trim={12cm 0cm 15cm 0cm},clip]{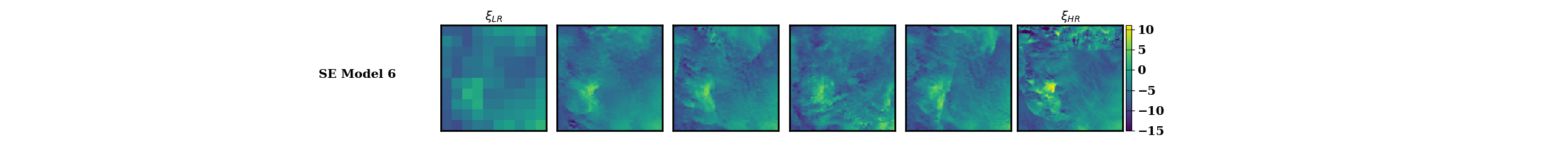}
    \includegraphics[width=0.97\textwidth,trim={12cm 0cm 15cm 0cm},clip]{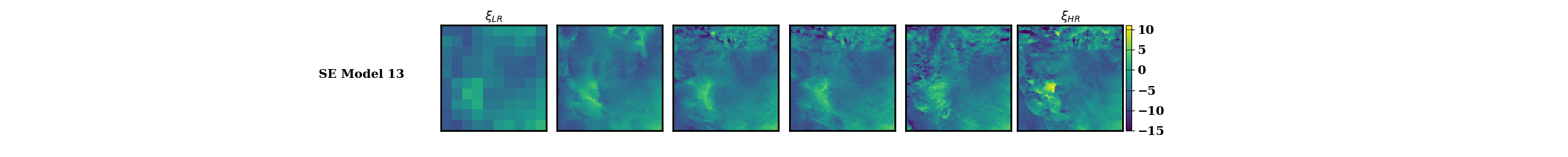}
    \includegraphics[width=0.97\textwidth,trim={12cm 0cm 15cm 0cm},clip]{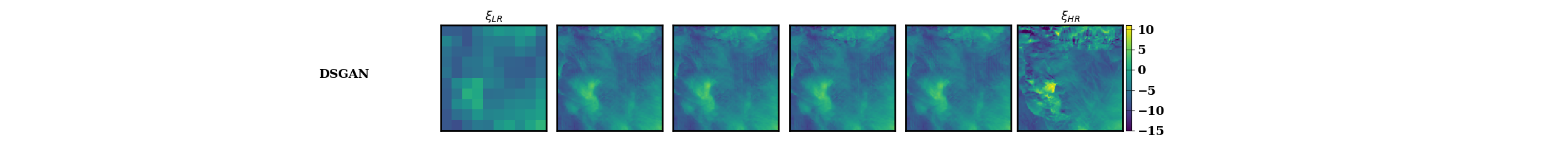}
    \includegraphics[width=0.97\textwidth,trim={12cm 0cm 15cm 0cm},clip]{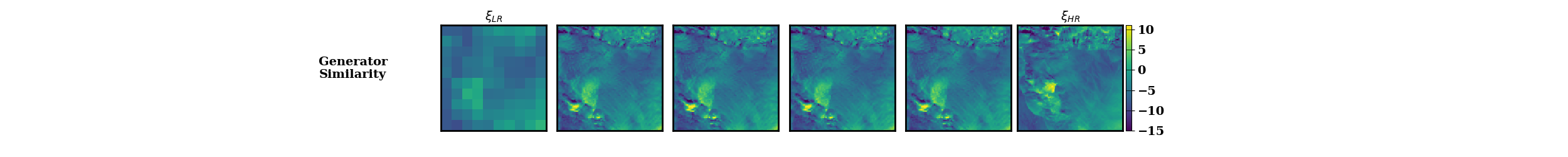}
    \caption{Contours of easterly wind velocity data. Left: LR wind velocity. Right: HR wind velocity. Middle four: examples of generated wind velocity conditioned on the same LR data.}
    \label{fig:illustrationDIV}
\end{figure}

The diversity of the generated samples is first assessed via visual examination of Fig.~\ref{fig:illustrationDIV}. For each model, a random vector $z$ is fed to the input of the generator on top of the LR field. It can be observed that for the methods that match a priori estimated moments, diverse velocity fields are generated (first three rows of Fig.~\ref{fig:illustrationDIV}. The variations between the fields are mostly localized near shear layers and in regions of energetic small scales (local high levels of turbulence). In contrast, for the DSGAN method (fourth row of Fig.~\ref{fig:illustrationDIV}) and the Generator Similarity method (sixth row of Fig.~\ref{fig:illustrationDIV}), little diversity can be observed. 

\begin{figure}[h!]
  \centering
  \includegraphics[width=0.97\textwidth,trim={0cm 0cm 0cm 0cm},clip]{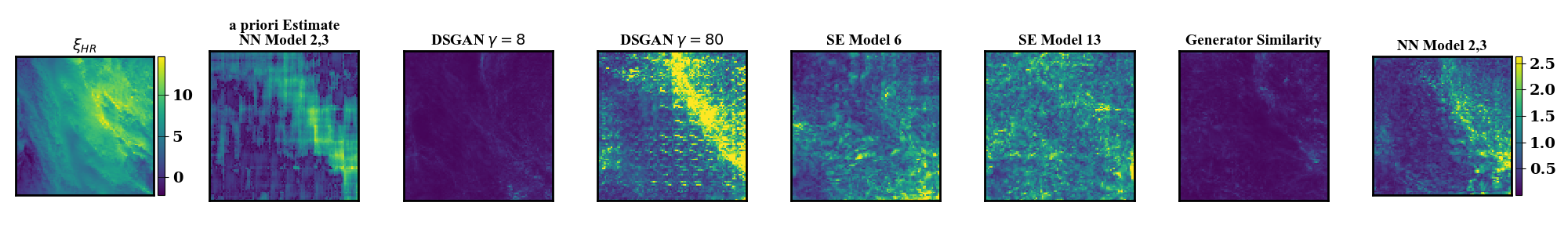}
    \caption{Contour of easterly wind velocity. Left: HR wind velocity. Next five right: conditional diversity estimated ($\sigma(X_{SF}|X_{LR})$) for the models compared.}
    \label{fig:Conditionalstatistics}
\end{figure}

Conditional standard deviations of the generated samples, conditioned on the LR data are shown in Fig.~\ref{fig:Conditionalstatistics}. It can be seen that the conditional diversity of samples generated with the Generator Similarity and DSGAN are much smaller than the a priori estimate of the diversity. In contrast, for the methods proposed, the amount of diversity between the generated fields and the a priori estimate is similar. Additionally, a large amount of diversity is generated in regions of large shear layers. It can also be observed that the generated fields exhibit diversity where none is predicted by the a priori estimate. Since this effect is most pronounced for the stochastic estimates of the conditional moments, it is postulated that the difference in the distribution of diversity is due to the estimate of the conditional moments used in the diversity loss.  

\subsubsection{Quantitative comparison of models}

The qualitative observations, along with other properties of the generated fields are assessed quantitatively in this section. The statistics shown in Table~\ref{tab:results} are obtained from the validation dataset (not shown during the training). When applicable, the uncertainty ranges correspond to the statistical uncertainty when computing expected values of the type $\mathbb{E}_{\bar{\xi}}$. 

\begin{table}[t]
\begin{center}
\setlength{\abovecaptionskip}{0pt}
\setlength{\belowcaptionskip}{5pt}
\caption{Quantitative comparison of the model consistency (second column), diversity (third and fourth column), and precision (fifth column).}
\label{tab:results}
\begin{tabular}{lcccc}
\hline
%Model  & Mismatch $\bar{\xi}$ Eq.~\ref{eq:contentMetric} [\%]   &  Diversity Metric Eq.~\ref{eq:divMetric} [\%]  & FID $U^{(1)}$~/~FID $U^{(2)}$ &   F_{1/8} $U^{(1)}$~/~F_{1/8} $U^{(2)}$   \\ \hline
Model  & Consistency metric [\%]   &  Diversity Metric [\%]  & FID $U^{(1)}$~/~FID $U^{(2)}$ &   F_{1/8} $U^{(1)}$~/~F_{1/8} $U^{(2)}$   \\ \hline
DSGAN & $4.40 \pm 0.02$  & $89 \pm 1$ & $\boldsymbol{53.15~/~51.43}$ & $\boldsymbol{0.955~/~0.955}$   \\
%DSGAN $\gamma=80$ & $2.26 \pm 0.01$  & $90 \pm 3$ & $171.55~/~152.53$ & $0.256~/~0.236$  \\
Generator Similarity & $4.39 \pm 0.02$  & $88 \pm 1$ & $\boldsymbol{51.08~/~53.00}$ & $\boldsymbol{0.975~/~0.944}$  \\
\textbf{SE Model 6} & $7.61 \pm 0.05$  & $\boldsymbol{62 \pm 1}$ & $96.39~/~111.31$ & $0.605~/~0.543$  \\
\textbf{SE Model 13} & $7.23 \pm 0.05$  & $\boldsymbol{61 \pm 2}$ & $100.35~/~100.92$ & $0.574~/~0.589$  \\
\textbf{NN Model 2,3} & $6.10 \pm 0.04$ & $\boldsymbol{43 \pm 1}$ & $103.77~/~113.99$ & $0.578~/~0.533$  \\
ADM n=5 & $\boldsymbol{0.78 \pm 0.02}$ & NA & $201.63~/~188.80$ & $0.017~/0.108$  \\
Taylor n=2 & $\boldsymbol{2.55 \pm 0.07}$ & NA & $222.80~/~208.45$ & $0.024~/~0.116$  \\

\hline

\end{tabular}
\end{center}
\vspace{0mm} 
\end{table}

To assess the level of diversity in the samples, the diversity metric (Eq.~\ref{eq:divMetric}) is computed for all the models except ADM and Taylor expansion, which produce a single output. The a priori estimate of the moments is taken to be obtained from Model 2 of the NN-assisted estimation method (see Sec.~\ref{sec:srwindNNEstimation}). The best performances for diversity are achieved for the three models proposed (highlighted in bold in Tab.~\ref{tab:results}). Interestingly, although the model implemented with stochastic estimation used suboptimal estimates of the conditional moments, they still achieved a better performance than DSGAN and the Generator Similarity technique. This suggests, that even suboptimal estimates of the first and second conditional moments still produce reasonable results. 

For the evaluation of the quality of samples, a pointwise comparison between the generated samples and the HR samples is not appropriate. Since the objective is to generate all the possible $\xi$ that map to $\bar{\xi}$, it is not expected that all the generated samples will be close to the HR observation. Instead, several metrics have been suggested in the image processing community, and are commonly used for the evaluation of samples generated by GANs. Here, the Fr\'{e}chet Inception distance (FID) \cite{heusel2017gans} and the $F_{1/8}$ score \cite{sajjadi2018assessing} are implemented and are briefly described. For the FID, the images generated are processed by one of the layers of the Inception network \cite{szegedy2015going} to obtain a latent representation of the images. The distributions of the latent space variables between the true data and the generated data are compared via the Fr\'{e}chet distance. An issue of the FID is that it assesses at the same time diversity and quality of the samples. To disentangle quality and diversity, precision and recall scores were developed \cite{sajjadi2018assessing,kynkaanniemi2019improved}. The precision, via the $F_{1/8}$ score is evaluated by estimating how much of the generated data is part of the support of the true data. This estimation is also performed using the Inception network so that the support of the true and generated distribution are estimated based on the features of the data. 

Since the Inception network was trained to process images, the wind data must be transformed into RGB data to compute the FID and $F_{1/8}$ scores. The easterly velocity $U^{(1)}$ and northerly velocities $U^{(2)}$ are transformed into two sets of grey scales images where pixels values are integers rescaled between $0$ and $255$. The FID and $F_{1/8}$ scores are then computed separately for each velocity component, and the results are shown in the last two columns of Tab.~\ref{tab:results}. The models with the least amount of diversity generate the images with the highest precision. Although the models proposed do not generate data with the highest precision, they do maintain high-fidelity spatial statistics and are the only ones to provide a good compromise between precision and diversity.

Finally, the mismatch between the filtered generated field $g(\widehat{\xi})$ and the target filtered field $\bar{\xi}$ is quantified using Eq.~\ref{eq:contentMetric}, and the results are shown in Tab.~\ref{tab:results} (first column). While small deviations may be observed (at most $7.61 \%$ of $||\bar{\xi}||_2$), the generated data is consistent with $\bar{\xi}$ for all the methods. This result suggests that the content loss is a reasonable technique to enforce $g(\widehat{\xi}) = \bar{\xi}$.

\section{Advantages and disadvantages}
\label{sec:advantages}

The method presented here can be used for any type of data as long as the conditional moments vary smoothly with the conditional variables. In particular, nothing prevents the same method to be used with data that is not structured. The estimation of the conditional moments can be done offline and does not impact the main computational cost which is the training process. This means that more advanced and costly techniques for the estimation of the conditional moments could be used while only marginally affecting the overall cost. For ease of implementation, the conditional moments were all precomputed and stored as part of a new dataset. In terms of memory requirements, the new diversity loss requires a dataset about three times larger: each training sample is associated with two additional fields (the first and the second moments). In the future, a more efficient approach could be developed where instead of storing the conditional moments, one could call the polynomial functions (in the stochastic estimation cases) or the external neural network (in the case of neural network-assisted estimation) to compute estimated moments on-the-fly. It was observed that the new diversity loss led to faster training than DSGAN and Generator Similarity techniques (about 1.5 times faster per epoch). This finding is in line with the work of Wu et al.~\cite{wu2020enforcing} where enforcing statistics for the generated data led to faster and more stable training. Finally, it can be expected that quality of the samples does not strongly depend on the filter size used. For example, non-conditional GANs are able to generate high-quality samples without conditioning on any input \cite{goodfellow2014generative}. In addition, it was found in a separate study that super-resolution could be achieved for atmospheric flows with filter size five times larger than the one considered here \cite{stengel2020adversarial}.

The main shortcoming of the formulation is the assumption that the covariance matrix of the unresolved data can be considered diagonal. From the results obtained, it appears that the discriminator successfully compensated for the approximation and it was still possible to generate high-quality results. Additionally, a finer description of the distribution can be implemented by assuming the covariance matrix to be diagonal-by-block. In that case, more moments would need to be computed during a priori analysis. A second issue in the method is that there is no guarantee that an optimal estimate of the conditional moments has been found. To the authors' knowledge, the only way to minimize the estimate error is to gradually increase the complexity of the conditional moment estimator until no more improvement is observed. Albeit costly, this approach was shown to be tractable even in the case of high-dimensional $\xi$ and $\bar{\xi}$.

\section{Conclusions}
\label{sec:conclusions}

In this work, it was demonstrated how an adversarial approach can be used to sample conditional high-dimensional distributions, which is critical in many physics problems. When continuous conditioning variables are used, even if no two samples share the same conditioning variable value, conditional moments can be estimated. Conditional generative models can be evaluated against the estimated conditional moments. It is also shown that by incorporating conditional moments as part of the loss function, it is possible to improve the diversity of the generated samples.

The method was demonstrated for the deconvolution of turbulent atmospheric flow data. GAN-based methods produced significantly higher fidelity samples than other methods. The method proposed here only marginally decreased the quality of the generated samples while significantly increasing their diversity. In particular, no matter the method used for the estimation of conditional moments, the diversity of the generated samples was found to be larger than other methods which encourage diversity without telling the generator where diversity should be generated. Finally, it was found that all the methods investigated, generated deconvolved fields consistent with their convolved counterparts.

Several improvements could be explored as future work. The conditional moments are affected by modeling and statistical uncertainty. Including these uncertainties in the loss function would avoid forcing the generation of diverse samples solely because of errors in the moment estimation. For the wind data super-resolution, the moments of pixel values were evaluated but the training could be accelerated by first conducting a modal decomposition of the samples and computing the conditional moments of a reduced number of modes. Additionally, the conditional moments are obtained with neural net architecture or a functional basis that is deemed best for all the snapshots. Different neural net architecture or functional basis may likely be adequate for different snapshots. Therefore, it could be advantageous to first separate the data into different classes and then compute a different model for each class. Furthermore, the main assumption of a Gaussian field with a diagonal covariance matrix could be relaxed with a diagonal-by-block covariance matrix.

As future work, it will be useful to investigate how transferable are the learned moments. In a practical implementations, the conditional moments will likely be learned with data gathered on a surrogate of the real device. The moments may need to be adapted to take advantage, for example, of scale similarity between the real and the surrogate device.

\section*{Acknowledgements}
This work was authored by the National Renewable Energy Laboratory (NREL), operated by Alliance for Sustainable Energy, LLC, for the U.S. Department of Energy (DOE) under Contract No. DE-AC36-08GO28308. This work was supported by the Laboratory Directed Research and Development (LDRD) Program at NREL and by funding from DOE's Advanced Scientific Computing Research (ASCR) program. The research was performed using computational resources sponsored by the Department of Energy's Office of Energy Efficiency and Renewable Energy and located at the National Renewable Energy Laboratory. The views expressed in the article do not necessarily represent the views of the DOE or the U.S. Government. The U.S. Government retains and the publisher, by accepting the article for publication, acknowledges that the U.S. Government retains a nonexclusive, paid-up, irrevocable, worldwide license to publish or reproduce the published form of this work, or allow others to do so, for U.S. Government purposes.

\addcontentsline{toc}{section}{Appendices}
\section*{Appendix}
\renewcommand{\thesubsection}{\Alph{subsection}}

\subsection{Loss balancing}
\label{app:lossBalance}
\begin{figure}[ht!]
\centering
\includegraphics[width=0.45\textwidth,trim={0cm 0cm 0cm 0cm},clip]{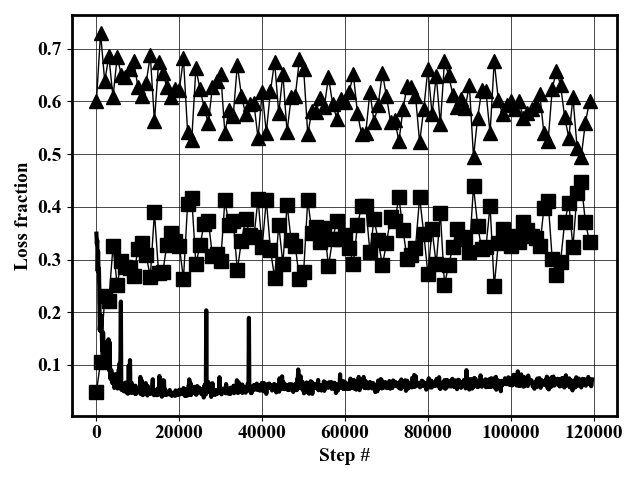}
\caption{Relative contribution of content loss (\mythickline{black}), adversarial loss (\mythickbarredsquare{black}{black}) and diversity loss (\mythickbarredtriangle{black}{black}) in the loss function of the generator trained with neural network assisted moment estimation at every training step.}
\label{fig:lossBalance}
\end{figure}

The coefficients $\alpha$ $\beta$ and $\gamma$ control the relative importance of each term in the global loss function, and dictate what is the primary focus of the GAN. For example, if $\alpha$ is too small, the box-filtered super-resolved fields will depart from the input $\xi_{LR}$. Here, an equal importance for the content, adversarial and diversity losses is desired. Therefore, each term should be appropriately balanced so that it equally affects the global loss function. In other studies, loss balancing was found to be critical for GANs in a variety of contexts \cite{malkiel2020mtadam,stengel2020adversarial}. The rationale adopted here is to ensure that throughout the training procedure, the content, adversarial and diversity losses magnitudes should be not be negligible. We have experimented with different set of weights and eventually selected the set described in Sec.~\ref{sec:enforcingConsistency}. The contribution of each term is plotted in Fig.~\ref{fig:lossBalance} for the generator trained with neural network assisted moment estimation. It can be seen that while the magnitude of each term varies throughout the training process, no one loss contributes negligibly, thereby ensuring proper balance. Without loss balancing, the adversarial loss would dominate the generator loss which would lead to a lack of consistency with $\xi_{LR}$ or a lack of diversity.

\subsection{Advantage of quadratic stochastic estimation}
\label{app:lseqse}
Using linear stochastic estimation (LSE), the conditional moment estimation was performed using the first neighborhood and the second neighborhood of each LR pixel (see Fig.~\ref{fig:trainingSE}). The result of the LSE should be compared to Model 6 and Model 13 (see Tab.\ref{tab:modelSE}), which are the best models found with a quadratic stochastic estimation that use the first and second neighborhood. The results are shown in Tab.~\ref{tab:lseqse}. It can be observed that the LSE is outperformed by the quadratic approximation in both cases. Therefore, to get closer to the neural network assisted estimation, a LSE alone would require using larger stencil neighborhood, thereby risking overfitting the polynomial expansion. However, it can be expected that even a suboptimal LSE will provide reasonable enough moment estimates to outperform DSGAN and Generator similarity. As shown in Tab.~\ref{tab:results} in the manuscript, Model 6, which as accurate as LSE using both stencil neighborhoods, already generates reasonably good diversity.

\begin{table}
\setcellgapes{5pt}
\makegapedcells
\caption{Comparison of linear and quadratic stochastic estimation.}
\vspace*{0mm}
\label{tab:lseqse}
\begin{center}
\begin{tabular}{lccc}
\hline
Neighborhood & Type of stochastic estimation  & MSE $p=1$  &   MSE $p=2$           \\ \hline
\multirowcell{2}{$1^{st}$} & Linear & 2.06 & 43.26 \\ \cline{2-4}
                           & Quadratic & 1.94 & 37.54 \\ \hline
\multirowcell{2}{$2^{nd}$} & Linear & 1.92 & 37.77 \\ \cline{2-4}
                           & Quadratic & 1.72 & 32.06 \\ \hline                          
\end{tabular}
\end{center}
\vspace{0mm} 
\end{table}

%\section{Conclusions and future work}
%\label{sec:conclusions}

%\begin{itemize}
%    \item Get uncertainty on spatial or temporal super res
%    \item Generate proper distribution, we use that for generation of extreme events
%    \item Cluster images to create different stochastic estimation models
%    \item incorporate sampling noise and model uncertainty \cite{richardson2018gans}
%    \item Would other decomposition help? POD ? Autoencoder
%    \item extend this to other fluid, with tougher constraints, like turbulent combustion
%    \item we still have large difference for the second moments
%\end{itemize}

%\subsection{Alternatives}
%
%\begin{itemize}
%    \item Could we do it for the latent representation? Lower dimensions but: Spatial correlation are not as clear. Difficult to evaluate physically.
%    \item Why do we do it pixelwise: because we can handle inhomogeneity due to either hidden inhomogeneous variables or the filtering operation
%    \item Why do you not use a NN? Using a NN would make sense with a lot of data. Here, since we want to do it pixel wise, it would be a bit dangerous (not enough data)
%\end{itemize}

%\begin{itemize}
%    \item Future work
%    \begin{itemize}
%        \item INtegrate it in algorithm that requires sampling
%        \item Evaluate the applicability to unstructured conditional variable 
%    \end{itemize}
%\end{itemize}

%\bibliography{master.bib}

\begin{thebibliography}{10}
\expandafter\ifx\csname url\endcsname\relax
  \def\url#1{\texttt{#1}}\fi
\expandafter\ifx\csname urlprefix\endcsname\relax\def\urlprefix{URL }\fi
\expandafter\ifx\csname href\endcsname\relax
  \def\href#1#2{#2} \def\path#1{#1}\fi

\bibitem{daley1993atmospheric}
R.~Daley, Atmospheric data analysis, no.~2, Cambridge university press, 1993.

\bibitem{lorenz1963deterministic}
E.~N. Lorenz, Deterministic nonperiodic flow, Journal of the atmospheric
  sciences 20~(2) (1963) 130--141.

\bibitem{kalnay2003atmospheric}
E.~Kalnay, Atmospheric modeling, data assimilation and predictability,
  Cambridge university press, 2003.

\bibitem{hassanaly2019lyapunov}
M.~Hassanaly, V.~Raman, Lyapunov spectrum of forced homogeneous isotropic
  turbulent flows, Physical Review Fluids 4~(11) (2019) 114608.

\bibitem{hassanaly2019ensemble}
M.~Hassanaly, V.~Raman, {Ensemble-LES analysis of perturbation response of
  turbulent partially-premixed flames}, Proceedings of the Combustion Institute
  37~(2) (2019) 2249--2257.

\bibitem{toth1997ensemble}
Z.~Toth, E.~Kalnay, {Ensemble forecasting at NCEP and the breeding method},
  Monthly Weather Review 125~(12) (1997) 3297--3319.

\bibitem{buizza1995singular}
R.~Buizza, T.~N. Palmer, The singular-vector structure of the atmospheric
  global circulation, Journal of the Atmospheric Sciences 52~(9) (1995)
  1434--1456.

\bibitem{leutbecher2008ensemble}
M.~Leutbecher, T.~N. Palmer, Ensemble forecasting, Journal of computational
  physics 227~(7) (2008) 3515--3539.

\bibitem{hassanaly2021classification}
M.~Hassanaly, V.~Raman, Classification and computation of extreme events in
  turbulent combustion, Progress in Energy and Combustion Science 87 (2021)
  100955.

\bibitem{hassanaly2020data}
M.~Hassanaly, Y.~Tang, S.~Barwey, V.~Raman, {Data-driven Analysis of Relight
  variability of Jet Fuels induced by Turbulence}, Combustion and Flame 225
  (2020) 453--467.

\bibitem{ebi2016experimental}
D.~Ebi, N.~T. Clemens, Experimental investigation of upstream flame propagation
  during boundary layer flashback of swirl flames, Combustion and Flame 168
  (2016) 39--52.

\bibitem{koo2012large}
H.~Koo, V.~Raman, Large-eddy simulation of a supersonic inlet-isolator, AIAA
  journal 50~(7) (2012) 1596--1613.

\bibitem{morio2014survey}
J.~Morio, M.~Balesdent, D.~Jacquemart, C.~Verg{\'e}, A survey of rare event
  simulation methods for static input--output models, Simulation Modelling
  Practice and Theory 49 (2014) 287--304.

\bibitem{del2005genealogical}
P.~Del~Moral, J.~Garnier, et~al., Genealogical particle analysis of rare
  events, The Annals of Applied Probability 15~(4) (2005) 2496--2534.

\bibitem{cerou2007adaptive}
F.~C{\'e}rou, A.~Guyader, Adaptive multilevel splitting for rare event
  analysis, Stochastic Analysis and Applications 25~(2) (2007) 417--443.

\bibitem{wouters2016rare}
J.~Wouters, F.~Bouchet, Rare event computation in deterministic chaotic systems
  using genealogical particle analysis, Journal of Physics A: Mathematical and
  Theoretical 49~(37) (2016) 374002.

\bibitem{hassanaly2019self}
M.~Hassanaly, V.~Raman, A self-similarity principle for the computation of rare
  event probability, Journal of Physics A: Mathematical and Theoretical 52~(49)
  (2019) 495701.

\bibitem{fukami2020machine}
K.~Fukami, K.~Fukagata, K.~Taira, Machine learning based spatio-temporal super
  resolution reconstruction of turbulent flows, arXiv preprint arXiv:2004.11566
  (2020).

\bibitem{barwey2019using}
S.~Barwey, M.~Hassanaly, V.~Raman, A.~Steinberg, {Using machine learning to
  construct velocity fields from OH-PLIF images}, Combustion Science and
  Technology (2019) 1--24.

\bibitem{barwey2020extracting}
S.~Barwey, V.~Raman, A.~Steinberg, {Extracting Information Overlap in
  Simultaneous OH-PLIF and PIV Fields with Neural Networks}, arXiv preprint
  arXiv:2003.03662 (2020).

\bibitem{langford1999optimal}
J.~A. Langford, R.~D. Moser, {Optimal LES formulations for isotropic
  turbulence}, Journal of fluid mechanics 398 (1999) 321--346.

\bibitem{adrian1990stochastic}
R.~J. Adrian, Stochastic estimation of sub-grid scale motions, Applied
  Mechanics Review 43~(5) (1990).

\bibitem{pope2010self}
S.~B. Pope, Self-conditioned fields for large-eddy simulations of turbulent
  flows, Journal of Fluid Mechanics 652 (2010) 139.

\bibitem{meneveau1994statistics}
C.~Meneveau, {Statistics of turbulence subgrid-scale stresses: Necessary
  conditions and experimental tests}, Physics of Fluids 6~(2) (1994) 815--833.

\bibitem{xie2017approximate}
X.~Xie, D.~Wells, Z.~Wang, T.~Iliescu, Approximate deconvolution reduced order
  modeling, Computer Methods in Applied Mechanics and Engineering 313 (2017)
  512--534.

\bibitem{sagaut2006large}
P.~Sagaut, Large eddy simulation for incompressible flows: an introduction,
  Springer Science \& Business Media, 2006.

\bibitem{germano2009new}
M.~Germano, A new deconvolution method for large eddy simulation, Physics of
  Fluids 21~(4) (2009) 045107.

\bibitem{adams2004implicit}
N.~Adams, S.~Hickel, S.~Franz, Implicit subgrid-scale modeling by adaptive
  deconvolution, Journal of Computational Physics 200~(2) (2004) 412--431.

\bibitem{echekki2001one}
T.~Echekki, A.~R. Kerstein, T.~D. Dreeben, J.-Y. Chen, ``one-dimensional
  turbulence" simulation of turbulent jet diffusion flames: model formulation
  and illustrative applications, Combustion and Flame 125~(3) (2001)
  1083--1105.

\bibitem{stolz1999approximate}
S.~Stolz, N.~A. Adams, An approximate deconvolution procedure for large-eddy
  simulation, Physics of Fluids 11~(7) (1999) 1699--1701.

\bibitem{stolz2001approximate}
S.~Stolz, N.~A. Adams, L.~Kleiser, An approximate deconvolution model for
  large-eddy simulation with application to incompressible wall-bounded flows,
  Physics of fluids 13~(4) (2001) 997--1015.

\bibitem{domaradzki1997subgrid}
J.~A. Domaradzki, E.~M. Saiki, A subgrid-scale model based on the estimation of
  unresolved scales of turbulence, Physics of Fluids 9~(7) (1997) 2148--2164.

\bibitem{domingo2015large}
P.~Domingo, L.~Vervisch, Large eddy simulation of premixed turbulent combustion
  using approximate deconvolution and explicit flame filtering, Proceedings of
  the Combustion Institute 35~(2) (2015) 1349--1357.

\bibitem{domingo2017dns}
P.~Domingo, L.~Vervisch, {DNS and approximate deconvolution as a tool to
  analyse one-dimensional filtered flame sub-grid scale modelling}, Combustion
  and Flame 177 (2017) 109--122.

\bibitem{geurts1997inverse}
B.~J. Geurts, Inverse modeling for large-eddy simulation, Physics of Fluids
  9~(12) (1997) 3585--3587.

\bibitem{wang2019regularized}
Q.~Wang, M.~Ihme, A regularized deconvolution method for turbulent closure
  modeling in implicitly filtered large-eddy simulation, Combustion and Flame
  204 (2019) 341--355.

\bibitem{wang2019regularizedspray}
Q.~Wang, X.~Zhao, M.~Ihme, A regularized deconvolution model for sub-grid
  dispersion in large eddy simulation of turbulent spray flames, Combustion and
  Flame 207 (2019) 89--100.

\bibitem{akram2020priori}
M.~Akram, M.~Hassanaly, V.~Raman, A priori analysis of reduced description of
  dynamical systems using approximate inertial manifolds, Journal of
  Computational Physics 409 (2020) 109344.

\bibitem{maulik2018data}
R.~Maulik, O.~San, A.~Rasheed, P.~Vedula, {Data-driven deconvolution for large
  eddy simulations of Kraichnan turbulence}, Physics of Fluids 30~(12) (2018)
  125109.

\bibitem{nikolaou2018modelling}
Z.~M. Nikolaou, C.~Chrysostomou, L.~Vervisch, S.~Cant, Modelling turbulent
  premixed flames using convolutional neural networks: application to sub-grid
  scale variance and filtered reaction rate, arXiv preprint arXiv:1810.07944
  (2018).

\bibitem{yuan2020deconvolutional}
Z.~Yuan, C.~Xie, J.~Wang, Deconvolutional artificial neural network models for
  large eddy simulation of turbulence, Physics of Fluids 32~(11) (2020) 115106.

\bibitem{brunton2020machine}
S.~L. Brunton, B.~R. Noack, P.~Koumoutsakos, Machine learning for fluid
  mechanics, Annual Review of Fluid Mechanics 52 (2020) 477--508.

\bibitem{duraisamy2020machine}
K.~Duraisamy, {Machine learning-augmented Reynolds-averaged and Large Eddy
  Simulation Models of turbulence}, arXiv preprint arXiv:2009.10675 (2020).

\bibitem{fukami2018super}
K.~Fukami, K.~Fukagata, K.~Taira, Super-resolution reconstruction of turbulent
  flows with machine learning, arXiv preprint arXiv:1811.11328 (2018).

\bibitem{fukami2019super}
K.~Fukami, K.~Fukagata, K.~Taira, Super-resolution analysis with machine
  learning for low-resolution flow data, in: 11th International Symposium on
  Turbulence and Shear Flow Phenomena, TSFP 2019, 2019.

\bibitem{wang2020physics}
C.~Wang, E.~Bentivegna, W.~Zhou, L.~Klein, B.~Elmegreen, {Physics-Informed
  Neural Network Super Resolution for Advection-Diffusion Models}, arXiv
  preprint arXiv:2011.02519 (2020).

\bibitem{stengel2020adversarial}
K.~Stengel, A.~Glaws, D.~Hettinger, R.~N. King, Adversarial super-resolution of
  climatological wind and solar data, Proceedings of the National Academy of
  Sciences 117~(29) (2020) 16805--16815.

\bibitem{goodfellow2014generative}
I.~Goodfellow, J.~Pouget-Abadie, M.~Mirza, B.~Xu, D.~Warde-Farley, S.~Ozair,
  A.~Courville, Y.~Bengio, Generative adversarial nets, in: Advances in neural
  information processing systems, 2014, pp. 2672--2680.

\bibitem{salimans2016improved}
T.~Salimans, I.~Goodfellow, W.~Zaremba, V.~Cheung, A.~Radford, X.~Chen,
  Improved techniques for training {GANs}, in: Advances in neural information
  processing systems, 2016, pp. 2234--2242.

\bibitem{isola2017image}
P.~Isola, J.-Y. Zhu, T.~Zhou, A.~A. Efros, Image-to-image translation with
  conditional adversarial networks, in: Proceedings of the IEEE conference on
  computer vision and pattern recognition, 2017, pp. 1125--1134.

\bibitem{craiu2014bayesian}
R.~V. Craiu, J.~S. Rosenthal, {Bayesian Computation Via Markov Chain Monte
  Carlo}, Annual Review of Statistics and Its Application 1 (2014) 179--201.

\bibitem{metropolis1953equation}
N.~Metropolis, A.~W. Rosenbluth, M.~N. Rosenbluth, A.~H. Teller, E.~Teller,
  Equation of state calculations by fast computing machines, The journal of
  chemical physics 21~(6) (1953) 1087--1092.

\bibitem{hastings1970monte}
W.~K. Hastings, {Monte Carlo sampling methods using Markov chains and their
  applications} (1970).

\bibitem{tanner1987calculation}
M.~A. Tanner, W.~H. Wong, The calculation of posterior distributions by data
  augmentation, Journal of the American statistical Association 82~(398) (1987)
  528--540.

\bibitem{ledig2017photo}
C.~Ledig, L.~Theis, F.~Husz{\'a}r, J.~Caballero, A.~Cunningham, A.~Acosta,
  A.~Aitken, A.~Tejani, J.~Totz, Z.~Wang, et~al., Photo-realistic single image
  super-resolution using a generative adversarial network, Proceedings of the
  IEEE conference on computer vision and pattern recognition (2017) 4681--4690.

\bibitem{karras2020analyzing}
T.~Karras, S.~Laine, M.~Aittala, J.~Hellsten, J.~Lehtinen, T.~Aila, {Analyzing
  and improving the image quality of StyleGAN}, in: Proceedings of the IEEE/CVF
  Conference on Computer Vision and Pattern Recognition, 2020, pp. 8110--8119.

\bibitem{farimani2017deep}
A.~B. Farimani, J.~Gomes, V.~S. Pande, Deep learning the physics of transport
  phenomena, arXiv preprint arXiv:1709.02432 (2017).

\bibitem{yang2020physics}
L.~Yang, D.~Zhang, G.~E. Karniadakis, Physics-informed generative adversarial
  networks for stochastic differential equations, SIAM Journal on Scientific
  Computing 42~(1) (2020) A292--A317.

\bibitem{king2018deep}
R.~King, O.~Hennigh, A.~Mohan, M.~Chertkov, {From deep to physics-informed
  learning of turbulence: Diagnostics}, arXiv preprint arXiv:1810.07785 (2018).

\bibitem{kingma2013auto}
D.~P. Kingma, M.~Welling, Auto-encoding variational bayes, arXiv preprint
  arXiv:1312.6114 (2013).

\bibitem{de2019deep}
M.~T.~H. de~Frahan, S.~Yellapantula, R.~King, M.~S. Day, R.~W. Grout, Deep
  learning for presumed probability density function models, Combustion and
  Flame 208 (2019) 436--450.

\bibitem{mirza2014conditional}
M.~Mirza, S.~Osindero, Conditional generative adversarial nets, arXiv preprint
  arXiv:1411.1784 (2014).

\bibitem{huang2017stacked}
X.~Huang, Y.~Li, O.~Poursaeed, J.~Hopcroft, S.~Belongie, Stacked generative
  adversarial networks, in: Proceedings of the IEEE conference on computer
  vision and pattern recognition, 2017, pp. 5077--5086.

\bibitem{kim2020unsupervised}
H.~Kim, J.~Kim, S.~Won, C.~Lee, Unsupervised deep learning for super-resolution
  reconstruction of turbulence, arXiv preprint arXiv:2007.15324 (2020).

\bibitem{subramaniam2020turbulence}
A.~Subramaniam, M.~L. Wong, R.~D. Borker, S.~Nimmagadda, S.~K. Lele,
  {Turbulence Enrichment using Physics-informed Generative Adversarial
  Networks}, arXiv (2020) arXiv--2003.

\bibitem{bode2021using}
M.~Bode, M.~Gauding, Z.~Lian, D.~Denker, M.~Davidovic, K.~Kleinheinz,
  J.~Jitsev, H.~Pitsch, Using physics-informed enhanced super-resolution
  generative adversarial networks for subfilter modeling in turbulent reactive
  flows, Proceedings of the Combustion Institute (2021).

\bibitem{borji2019pros}
A.~Borji, {Pros and cons of GAN evaluation measures}, Computer Vision and Image
  Understanding 179 (2019) 41--65.

\bibitem{breuleux2010unlearning}
O.~Breuleux, Y.~Bengio, P.~Vincent, Unlearning for better mixing, Universite de
  Montreal/DIRO (2010).

\bibitem{tolstikhin2017adagan}
I.~O. Tolstikhin, S.~Gelly, O.~Bousquet, C.-J. Simon-Gabriel, B.~Sch{\"o}lkopf,
  {AdaGAN}: Boosting generative models, in: Advances in Neural Information
  Processing Systems, 2017, pp. 5424--5433.

\bibitem{szegedy2015going}
C.~Szegedy, W.~Liu, Y.~Jia, P.~Sermanet, S.~Reed, D.~Anguelov, D.~Erhan,
  V.~Vanhoucke, A.~Rabinovich, Going deeper with convolutions, in: Proceedings
  of the IEEE conference on computer vision and pattern recognition, 2015, pp.
  1--9.

\bibitem{heusel2017gans}
M.~Heusel, H.~Ramsauer, T.~Unterthiner, B.~Nessler, S.~Hochreiter, {GANs
  trained by a two time-scale update rule converge to a local Nash
  equilibrium}, in: Advances in neural information processing systems, 2017,
  pp. 6626--6637.

\bibitem{zhao2016energy}
J.~Zhao, M.~Mathieu, Y.~LeCun, Energy-based generative adversarial network,
  arXiv preprint arXiv:1609.03126 (2016).

\bibitem{zhu2017toward}
J.-Y. Zhu, R.~Zhang, D.~Pathak, T.~Darrell, A.~A. Efros, O.~Wang, E.~Shechtman,
  Toward multimodal image-to-image translation, in: Advances in neural
  information processing systems, 2017, pp. 465--476.

\bibitem{yang2019diversity}
D.~Yang, S.~Hong, Y.~Jang, T.~Zhao, H.~Lee, Diversity-sensitive conditional
  generative adversarial networks, in: Proceedings of the International
  Conference on Learning Representations, 2019.

\bibitem{odena2018generator}
A.~Odena, J.~Buckman, C.~Olsson, T.~B. Brown, C.~Olah, C.~Raffel,
  I.~Goodfellow, {Is generator conditioning causally related to GAN
  performance?}, arXiv preprint arXiv:1802.08768 (2018).

\bibitem{arjovsky2017wasserstein}
M.~Arjovsky, S.~Chintala, L.~Bottou, {Wasserstein GAN}, arXiv preprint
  arXiv:1701.07875 (2017).

\bibitem{gulrajani2017improved}
I.~Gulrajani, F.~Ahmed, M.~Arjovsky, V.~Dumoulin, A.~C. Courville, {Improved
  training of Wasserstein GANs}, in: Advances in neural information processing
  systems, 2017, pp. 5767--5777.

\bibitem{metz2016unrolled}
L.~Metz, B.~Poole, D.~Pfau, J.~Sohl-Dickstein, Unrolled generative adversarial
  networks, arXiv preprint arXiv:1611.02163 (2016).

\bibitem{srivastava2017veegan}
A.~Srivastava, L.~Valkov, C.~Russell, M.~U. Gutmann, C.~Sutton, {Veegan:
  Reducing mode collapse in gans using implicit variational learning}, in:
  Advances in Neural Information Processing Systems, 2017, pp. 3308--3318.

\bibitem{wu2020enforcing}
J.-L. Wu, K.~Kashinath, A.~Albert, D.~Chirila, H.~Xiao, et~al., Enforcing
  statistical constraints in generative adversarial networks for modeling
  chaotic dynamical systems, Journal of Computational Physics 406 (2020)
  109209.

\bibitem{karras2017progressive}
T.~Karras, T.~Aila, S.~Laine, J.~Lehtinen, {Progressive growing of GANs for
  improved quality, stability, and variation}, arXiv preprint arXiv:1710.10196
  (2017).

\bibitem{papoulis2002probability}
A.~Papoulis, S.~U. Pillai, Probability, random variables, and stochastic
  processes, Tata McGraw-Hill Education, 2002.

\bibitem{adrian1989approximation}
R.~J. Adrian, B.~Jones, M.~Chung, Y.~Hassan, C.~Nithianandan, A.-C. Tung,
  Approximation of turbulent conditional averages by stochastic estimation,
  Physics of Fluids A: Fluid Dynamics 1~(6) (1989) 992--998.

\bibitem{adrian1994stochastic}
R.~J. Adrian, Stochastic estimation of conditional structure: a review, Applied
  scientific research 53~(3-4) (1994) 291--303.

\bibitem{dowson1982frechet}
D.~Dowson, B.~Landau, {The Fr{\'e}chet distance between multivariate normal
  distributions}, Journal of multivariate analysis 12~(3) (1982) 450--455.

\bibitem{shah2019encoding}
V.~Shah, A.~Joshi, S.~Ghosal, B.~Pokuri, S.~Sarkar, B.~Ganapathysubramanian,
  C.~Hegde, Encoding invariances in deep generative models, arXiv preprint
  arXiv:1906.01626 (2019).

\bibitem{sonderby2016amortised}
C.~K. S{\o}nderby, J.~Caballero, L.~Theis, W.~Shi, F.~Husz{\'a}r, Amortised map
  inference for image super-resolution, arXiv preprint arXiv:1610.04490 (2016).

\bibitem{draxl2015wind}
C.~Draxl, A.~Clifton, B.-M. Hodge, J.~McCaa, The wind integration national
  dataset (wind) toolkit, Applied Energy 151 (2015) 355--366.

\bibitem{abadi2016tensorflow}
M.~Abadi, P.~Barham, J.~Chen, Z.~Chen, A.~Davis, J.~Dean, M.~Devin,
  S.~Ghemawat, G.~Irving, M.~Isard, et~al., {Tensorflow: A system for
  large-scale machine learning}, in: 12th $\{$USENIX$\}$ Symposium on Operating
  Systems Design and Implementation ($\{$OSDI$\}$ 16), 2016, pp. 265--283.

\bibitem{pope2001turbulent}
S.~B. Pope, Turbulent flows (2001).

\bibitem{van1931einfluss}
P.~Van~Cittert, Zum einfluss der spaltbreite auf die intensit{\"a}tsverteilung
  in spektrallinien. ii, Zeitschrift f{\"u}r Physik 69~(5-6) (1931) 298--308.

\bibitem{san2015posteriori}
O.~San, A.~E. Staples, T.~Iliescu, A posteriori analysis of low-pass spatial
  filters for approximate deconvolution large eddy simulations of homogeneous
  incompressible flows, International Journal of Computational Fluid Dynamics
  29~(1) (2015) 40--66.

\bibitem{sajjadi2018assessing}
M.~S. Sajjadi, O.~Bachem, M.~Lucic, O.~Bousquet, S.~Gelly, Assessing generative
  models via precision and recall, in: Advances in Neural Information
  Processing Systems, 2018, pp. 5228--5237.

\bibitem{kynkaanniemi2019improved}
T.~Kynk{\"a}{\"a}nniemi, T.~Karras, S.~Laine, J.~Lehtinen, T.~Aila, Improved
  precision and recall metric for assessing generative models, in: Advances in
  Neural Information Processing Systems, 2019, pp. 3927--3936.

\bibitem{malkiel2020mtadam}
I.~Malkiel, L.~Wolf, {Mtadam: Automatic balancing of multiple training loss
  terms}, arXiv preprint arXiv:2006.14683 (2020).

\end{thebibliography}
%\bibliographystyle{elsarticle-num}

\end{document}